\RequirePackage{fix-cm}
\documentclass{svjour3}

\usepackage[morefloats=125]{morefloats}
\usepackage[hyphens]{url}
\usepackage{graphicx}
\usepackage{tabularx}
\usepackage{amssymb}
\usepackage{amsmath}
\usepackage{color}
\usepackage{wrapfig}
\usepackage{longtable}
\usepackage{morefloats}
\usepackage{float}
\usepackage{makecell}
\usepackage[breaklinks=true,hidelinks,pdfusetitle]{hyperref}
\usepackage{ifthen}
\usepackage{thmtools}
\usepackage{thm-restate}
\usepackage[sort&compress,square,comma,numbers]{natbib}
\usepackage{algorithm}
\usepackage[noend]{algpseudocode}
\usepackage[table,xcdraw]{xcolor}
\usepackage{multirow}
\usepackage{flexisym}
\usepackage{changepage}
\usepackage{subfig}
\usepackage{array}

\graphicspath{{./figures/}}

\newcounter{nullhyp}
\newenvironment{nullhyp}[1][]{\refstepcounter{nullhyp}\par\medskip
   \noindent \textit{H\thenullhyp0. #1} \rmfamily}

\renewcommand{\description}[1]{\vspace{\baselineskip}\noindent\textbf{#1}}

\newcommand{\ignore}[1]{}
\newcommand{\thickhline}{%
    \noalign {\ifnum 0=`}\fi \hrule height 1pt
    \futurelet \reserved@a \@xhline
}
\newlength{\Oldarrayrulewidth}

\newcolumntype{"}{@{\hskip\tabcolsep\vrule width 1pt\hskip\tabcolsep}}

\begin{document}

\title{Optimizing Prediction Intervals by Tuning Random Forest via Meta-Validation}

\author{
Sean Bayley \and
Davide Falessi
}

\institute{Sean Bayley \at
            Dept. of Computer Science and Engineering, Notre Dame\\
            \email{sbayley@nd.edu}
            \and
            Davide Falessi \at
              Dept. of Computer Science and Software Engineering, California Polytechnic State University\\
              \email{dfalessi@calpoly.edu}
}

\thispagestyle{plain}
\maketitle
\begin{abstract}
Recent studies have shown that tuning prediction models increases prediction accuracy and that Random Forest can be used to construct prediction intervals. However, to our best knowledge, no study has investigated the need to, and the manner in which one can, tune Random Forest for optimizing prediction intervals -- this paper aims to fill this gap. We explore a tuning approach that combines an effectively exhaustive search with a validation technique on a single Random Forest parameter. This paper  investigates which, out of eight validation techniques, are beneficial for tuning, i.e., which automatically choose a Random Forest configuration constructing prediction intervals that are reliable and with a smaller width than the default configuration. Additionally, we present and validate three meta-validation techniques to determine which are beneficial, i.e., those which automatically chose a beneficial validation technique. This study uses data from our industrial partner (Keymind Inc.) and the Tukutuku Research Project, related to post-release defect prediction and Web application effort estimation, respectively.  Results from our study indicate that: i) the default configuration is frequently unreliable, ii) most of the validation techniques, including previously successfully adopted ones such as 50/50 holdout and bootstrap, are counterproductive in most of the cases, and iii) the 75/25 holdout meta-validation technique is always beneficial; i.e., it avoids the likely counterproductive effects of validation techniques.
\end{abstract}







\section{Introduction}\label{sec:intro}

One important phase of software development is release planning \cite{ruhe2010product}. During release planning, stakeholders define characteristics of the software release such as the requirements to implement, the defects to fix, the developers to use, the amount and type of testing, and the release duration. The project manager must make decisions in the midst of conflicting goals like time-to-market, cost, number of expected defects, and customer demands. These business goals, their importance, and clients' expectations vary among releases of the same or different projects, and in turn the number of acceptable defects varies. For instance, some projects are more safety critical than others and, therefore, can tolerate fewer defects. Thus, the manager might decide to increase the amount of testing, reduce the number of features, or postpone the release deadline if the predicted number of defects does not fit the business goals of that project release.

In 2011, Keymind developed and institutionalized a tool to help managers define characteristics of a software release according to the predicted number of post-release defects. Over the last seven years, the tool was subject to eight major upgrades. Each upgrade included refreshing the data (i.e., collecting data about new software releases), refreshing the type of data (i.e., adding metrics based on changes in technologies being used), improving the usability of the layout, and selecting the most accurate prediction models. A major improvement effort took place in 2013, when Keymind transitioned from predicting the number of defects to predicting the upper bounds related to a confidence level \cite{falessi2014achieving}. Thus, Keymind moved from supporting a sentence such as ``The expected number of defects is $y$'' to supporting a sentence such as ``We are 90\% confident that the number of defects is less than $y$.'' Then, during a discussion with the Keymind team in 2016, we learned that the current model was actionable but not fully explicative. Specifically, we realized that an interval is more informative than its upper-bound, i.e., the lower bound is also informative. For instance, there is a practical difference in the intervals $0 \leq y \leq 10$ and $9 \leq y \leq 10$  constructed for two different software projects. On the one hand, the former is more desirable as the actual number of post-release defects could be 0, i.e., spending additional money on testing might not be required. On the other hand, the latter is more actionable, i.e., there is less variability. 
Upon reviewing the literature, we determined that prediction intervals can be used to support the sentence ``We are 90\% confident that the number of defects is between $a$ and $b$'', where $a$ and $b$ are the lower and upper bounds of the prediction interval \cite{jorgensen2003effort, angelis2000simulation, jorgensen2002combination}.


\subsection{Definitions}
In order to avoid ambiguities in the remainder of this paper, we define the following key terms and concepts.

\begin{itemize}
    \item \textbf{Model}: a formal description of structural patterns in data used to make predictions (e.g., Random Forest)  \cite{witten2016data}.
    \item \textbf{Parameter}: an internal variable of the algorithm used to learn the model (e.g., MTRY \cite{witten2016data}).
    \item \textbf{Configuration}: the set of values for each of the parameters of a model (e.g., MTRY=1.0 and all other parameters set as default).
    \item \textbf{Prediction Interval (PI)}: an estimate of an interval, with a certain probability, in which future observations will fall (e.g., $0 \leq \text{post-release defects} \leq 5$) \cite{geisser1993predictive}. An important difference between the confidence interval (CI) and the PI is that the PI refers to the uncertainty of an estimate, while the CI refers to the uncertainty associated with the parameters of a distribution \cite{jorgensen2003effort}, e.g., the uncertainty of the mean value of a distribution of values.
    \item \textbf{Width}: the size of a PI. Width is computed as the upper bound minus the lower bound, e.g., $Width(0 \leq y \leq 5) = 5$. The smaller the width, the better.
    \begin{figure}
        \centering
        \includegraphics[width=1.0\textwidth]{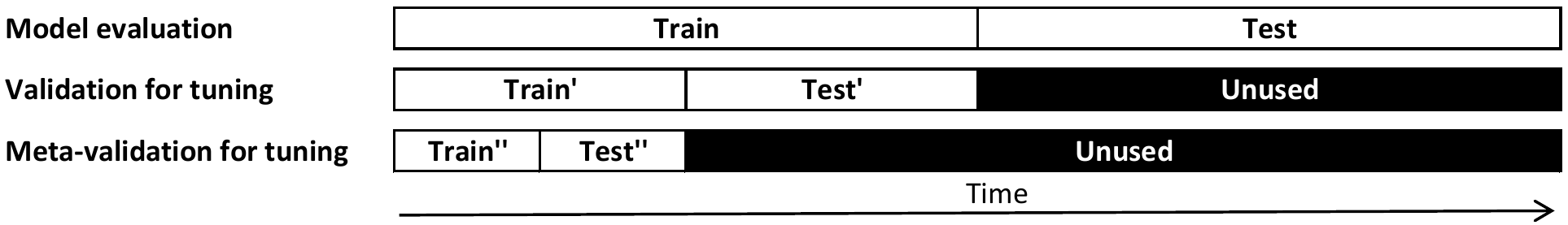}
        \caption{Data used for model evaluation, validation and meta-validation for tuning.}
        \label{fig:data}
    \end{figure}
    \item \textbf{Coverage}: also called coverage probability, hit-rate, or actual confidence, is the proportion of time the true value falls within the PI \cite{dodge2006oxford}. For instance, the true value might fall within the PI of a model 92\% of the time. The higher the coverage, the better.
    \item \textbf{Nominal Confidence (NC)}: the stated proportion of time the true value should fall within the PI \cite{dodge2006oxford}. A project manager might like to have a model that constructs PI that contain the true value 90\% of the times, i.e., the nominal confidence is 90\%.
    \item \textbf{Reliable}: refers to a model that constructs PIs such that $Coverage \geq NC$. For instance, a model is said to be reliable if nominal confidence is 90\% and coverage is 92\%. Reliable models are preferable.
    \item \textbf{Best Configuration}: with respect to PI, the goodness of a configuration is (1) contingent upon reliability and (2) inversely proportional to width. Thus, we define the best configuration as the configuration with smallest width among the set of reliable configurations.
    \item \textbf{Model evaluation}: the process of measuring the performance of a model, as configured in a specific way (i.e., default configuration). Typically the model is first trained (i.e., learned) and then it is tested (i.e., accuracy is measured).  For instance, in Figure~\ref{fig:data}, the dataset is first ordered chronologically and then split in two parts: train and test.  Note that the proportion of train and test (i.e., 66/33) is consistent in \citet{fu2016tuning} and \citet{tantithamthavorn2016automated}. We also note that there are validation techniques that split the dataset in proportions other than 66/33 or that do not even preserve the order of the data between training and testing. We discuss validation techniques in further detail in Section~\ref{sec:related_validation}.
    \item \textbf{Tuning, auto-configuration, or configurations validation}: is the process of automatically finding the model configuration which optimizes an objective function \cite{thornton2013auto, bergstra2011algorithms}. In our context, we are interested in selecting the configuration which constructs the narrowest reliable PI. In this work, because we let only one parameter vary, and because we evaluate all possible values of said parameter, a tuning technique coincides with an exhaustive search and a specific validation technique. Thus, in this work, the terms tuning technique and validation technique can be used interchangeably. Tuning uses the training set of the model evaluation for validating candidate model configurations. Specifically, in Figure~\ref{fig:data}, the evaluation training set is divided into \textit{train$^\prime$} and \textit{test$^\prime$} of tuning. We note that train and test sets of tuning can be of different proportions; the only constraint of a tuning technique is that it is not exposed to the evaluation test set. 
    \item \textbf{Beneficial:} refers to a tuning technique that is able to select a configuration that constructs PIs that are reliable and narrower than what the default configuration constructs.  
    \item \textbf{Configuration meta-validation}: aka meta-validation, is the process of automatically finding the most beneficial tuning  technique. In other words, if tuning selects the best configuration, by validating the different configurations, meta-validation selects the best technique to select the best configuration. Meta-validation uses the tuning training set for validating the tuning techniques. In Figure~\ref{fig:data}, the \textit{train$^\prime$} subset is further divided into \textit{train$^{\prime\prime}$} and \textit{test$^{\prime\prime}$}. Again, training and test sets can have different proportions; the only constraint is that no meta-validation subset use \textit{test} or \textit{test$^\prime$}.
\end{itemize}


\subsection{Motivation}\label{sec:motivation}

\begin{figure*}[!t]
    \centering
    \includegraphics[width=\textwidth,height=.4\paperheight,keepaspectratio]{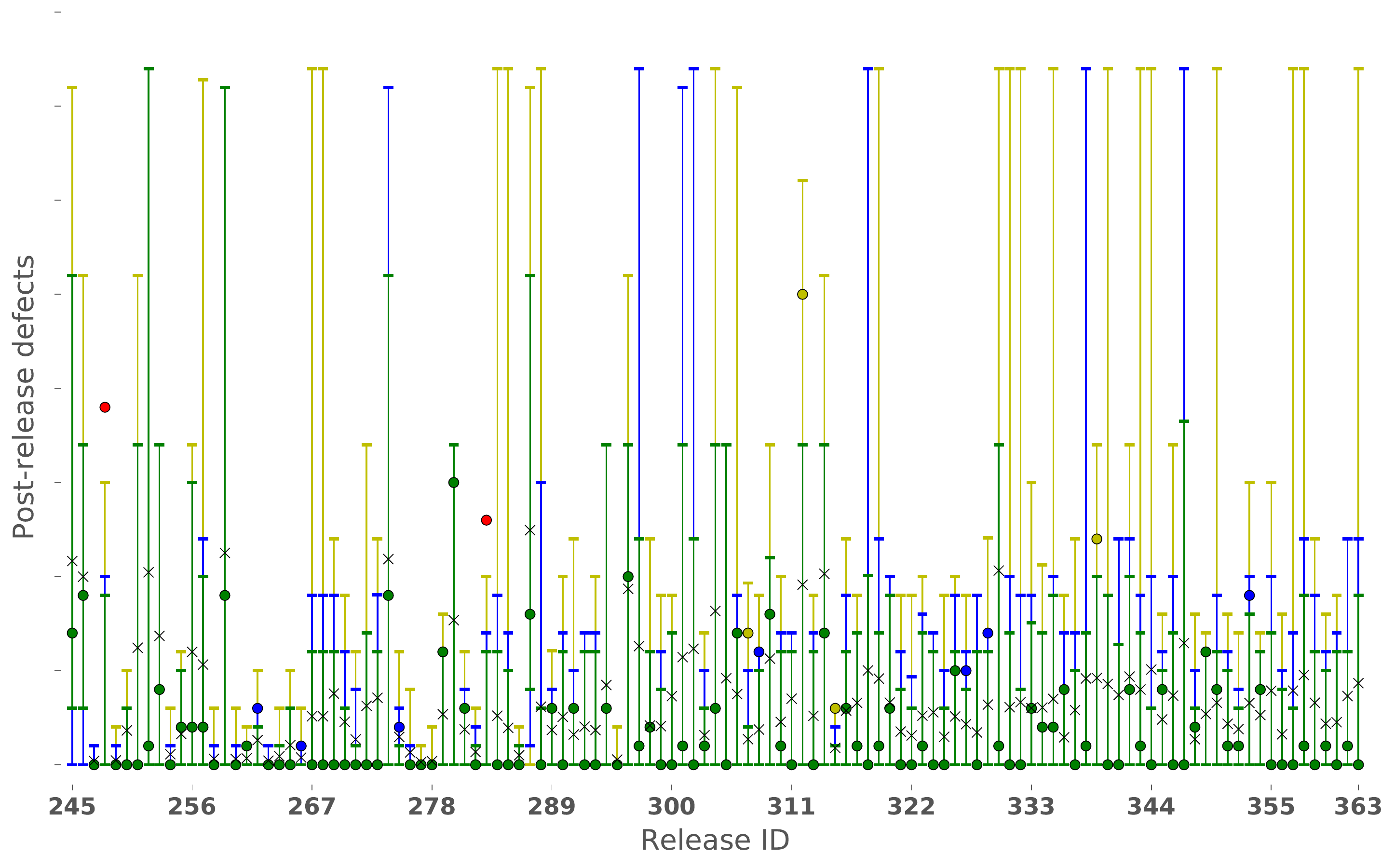}
    \caption{Prediction intervals at NC90 (green), NC95 (blue), and NC99 (yellow) as constructed by the default random forests configuration on the Keymind dataset. The model is trained on the first 66\% of the data and PIs are constructed for the remaining 33\%.}
    \label{fig:keymind_intervals}
\end{figure*}

In order to better understand the problem we are trying to solve, Figure~\ref{fig:keymind_intervals} reports PIs at NC90 (green), NC95 (blue), and NC99 (yellow) as constructed by the default random forest configuration on the Keymind dataset. The model is trained on the first 66\% of the data and PIs are constructed for the remaining 33\%. We plot the actual number of post-release defects with the same color of the narrowest interval in which it is contained, e.g., a green point indicates that the actual value is contained in the PI constructed at NC90. We plot points that are not contained at any NC as red. Lastly, we plot the predicted value with a ``x''. 

In Figure~\ref{fig:keymind_intervals}, we observe that NC90 covers 89\% (106), NC95 covers 95\% (113), and NC99 covers 98\% (117) of the total (119) releases. In other words, the default random forest configuration is not reliable at NC90 and NC99. Thus, we cannot use the default configuration to support the claim ``We are \textit{NC}\% confident that the number of defects is between \textit{a} and \textit{b}'' if \textit{NC} is 90 or 99. This motivates the need for tuning to choose a configuration, other than default, which constructs reliable PIs.

Figure~\ref{fig:keymind_intervals} is also useful to identify the differences between PIs and CIs. Specifically, there are cases in which the NC95 and NC99 intervals cannot be seen (e.g., release 252). This means that the intervals constructed at all NCs are equivalent. This denotes a significant difference between PIs and CIs: the width of CIs is strongly correlated with NC. Moreover, within the same NC, higher upper bounds do not always correspond with more predicted defects. For example, release 280 has a higher number of predicted defects than release 267 but a lower upper bound at NC95 and NC99. This denotes another significant difference between PIs and CIs. Specifically, within the same NC, the upper bound of CIs is strongly correlated with the predicted number (i.e., NC50).

Finally, it is interesting to note that larger intervals do not always correspond with more actual defects. For example, release 248 has the second most defects and the 27th, 33rd, and 47th largest interval for NC90, NC95, and NC99, respectively.


\subsection{Aim}

Recent studies have shown that tuning prediction models
increases accuracy \cite{duan2003evaluation, di2011genetic, corazza2013using, borg2016tuner, fu2016tuning}, and that Random Forest can be used to to predict defect and effort data \cite{Moeyersoms:2015, DBLP:journals/iee/SatapathyAR16} and to construct prediction intervals. However, to our best knowledge, no study has investigated the need, or how, to tune Random Forest for optimizing prediction intervals; this paper aims to fill this gap.

The contribution of this paper is threefold;
\begin{enumerate}
    \item We use Random Forest to construct prediction intervals for software engineering data (fault and effort).
    \item We present and evaluate the use of eight validation techniques for tuning to determine which are beneficial.
    \item We present and evaluate three meta-validation techniques to determine which are beneficial, i.e., which automatically choose a beneficial validation technique.
\end{enumerate}
Specifically, we investigate the following research questions.
\begin{restatable}{RQ}{rqtunepromise}\label{rq:tune_promise}
Is default the best configuration?
\end{restatable}
\vspace{-.9\baselineskip}
\begin{restatable}{RQ}{rqtunebenefit}\label{rq:tune_benefit}
Are validation techniques beneficial?
\end{restatable}
\vspace{-.9\baselineskip}
\begin{restatable}{RQ}{rqmetatunebenefit}\label{rq:metatune_benefit}
Are meta-validation techniques beneficial?
\end{restatable}

Our validation uses data from our industrial partner (Keymind Inc.) and the Tukutuku Research Project, containing 363 and 195 industrial data points related to post-release defect prediction and Web application effort estimation, respectively. 

Results show that the default configuration is frequently unreliable  and previously successfully adopted validation techniques, such as 50/50 holdout and bootstrap, are frequently counterproductive in tuning. Moreover, no \textit{single} validation technique is always beneficial; however, the \textit{meta} 75/25 holdout technique selects validation techniques that are always beneficial. Thus, results show that Random Forest is a viable solution to construct prediction intervals only if well tuned. Because no single validation technique resulted always beneficial, we recommend the use of \textit{meta} 75/25 holdout meta-validation technique to dynamically chose the validation technique according to the dataset and prediction interval of interest.

In order to support the usability and replicability of this study, we provide a Python package for tuning and meta-tuning Random Forests PIs. We also show how to use it over a large open source project called Apache-ANT.


\subsection{Structure}
The remainder of the paper is structured as follows. Section~\ref{sec:related} discusses related work. Section~\ref{sec:design} describes our experimental design. Section~\ref{sec:results} present results and discussion. Section~\ref{sec:replicability} presents \textit{meta\_tune}, an open source Python package we developed for using and tuning prediction intervals. Section \ref{sec:threats} discusses threats to validity. Section~\ref{sec:conclusion} concludes the paper and identifies areas for future work.

\section{Related Work}\label{sec:related}

\subsection{Defect Prediction}\label{sec:related_defect}
Numerous recent studies have applied data mining and machine learning techniques to improve QA resource allocation, i.e., to focus QA efforts on artifact(s) expected to be the most defective. These studies can be generally categorized as defect classification (i.e., the artifact is either defective or not defective) \cite{kim2011dealing, rahman2013sample, herzig2013s, zhang2014towards, mishra2012defect, selvaraj2013support, kim2008classifying, aversano2007learning, zimmermann2007predicting, menzies2013local, okutan2014software, guo2004robust, menzies2010defect, minku2015make, turhan2009relative, menzies2011local, suffian2014prediction, giger2012method, padberg2002empirical} or defect prediction (i.e., the artifact contains \textit{n} defects) \cite{li2003selecting, petersson2004capture, thelin2002confidence, biffl2001evaluating, biffl2000using, thelin2004team}.

Similar to this paper, \citet{li2003selecting} investigate the use of post-release defect prediction in the planning of widely used multi-release commercial systems. They conclude that models currently available fall short because they do not adequately consider organizational changes and customer adoption characteristics.

A family of studies apply capture-recapture to defect prediction \cite{petersson2004capture, thelin2002confidence, biffl2001evaluating, thelin2004team, padberg2002empirical}. These studies  estimate the remaining number of defects by analyzing characteristics of the overlapping set of defects detected by multiple, independent reviewers. Capture-recapture was first applied to software inspections by \citet{eick1992estimating} in 1992 and \citet{petersson2004capture} provide a survey of the method. 


\subsection{Effort Prediction}\label{sec:related_effort}
It is often difficult to predict the amount of effort required to complete a software project due to changes associated with aspects of software development (e.g., requirements, development environments, and personnel). It is of practical importance to have accurate effort predictions as it affects project management and resource allocation, i.e., larger parts of the budget should be allocated for artifacts that are expected to require more effort. Software effort prediction has been the subject of numerous studies, as summarized by \citet{boehm2000software} and \citet{molokken2003review}.

\citet{jorgensen2002combination} use independent expert opinions to predict the amount of development effort. They suggest that it is better to have estimation teams comprising several different types of roles, rather than estimation teams consisting of only technical roles, to help reduce systematic bias. We incorporate this suggestion into the development of the Keymind dataset, which we discuss further in Section~\ref{sec:design_keymind}.

\citet{briand1998cobra} develop and apply CoBRA, a hybrid method which combines aspects of algorithmic and experiential approaches, to predicting the cost of software development.  To construct CoBRA, domain experts are asked to decide the causal factors  and their possible values. Experimental results show realistic uncertainty estimates. However, this method requires, and is sensitive to, human estimates.

\citet{di2011genetic} propose and validate the use of a Genetic Algorithm with a grid search for tuning support vector regression for estimating defective file. \citet{corazza2013using} propose and validate a meta-heuristics Tabu Search for tuning support vector regression for point values effort estimation. Their results show that Tabu search outperformed other  simpler approaches, such as random configuration and default configuration. Because in this work it is reasonable to tune only one parameter, then a full search is preferable to any other incomplete search such as the Genetic or Tabu types.

\citet{Whigham:2015:BMS:} propose a baseline for comparing effort estimation models. Unfortunately such a baseline does not apply to prediction interval and therefore cannot be used in this work.

To the best of our knowledge, there has been no study that has applied Random Forest to effort prediction.


\subsection{Intervals in Software Engineering}\label{sec:related_intervals}
Predictions about the number of post-release defects or development effort are intrinsically uncertain. This uncertainty primarily arises as a result of the nature of software development (e.g., changing requirements, unstable development teams and environments). However, there is also some degree of uncertainty associated with the model used to make the prediction. Hence, information about the level of uncertainty is desirable. 

Software engineering studies using intervals can be generally categorized as either using CIs in combination with capture-recapture methods \cite{padberg2002empirical, petersson2004capture, thelin2002confidence, thelin2004team}, or constructing PIs from expert opinion elicitation \cite{jorgensen2002combination}, bootstrapping \cite{angelis2000simulation}, regression models \cite{braga2007software}, a combination of 
cluster analysis and classification methods \cite{bakir2011comparative},a multi-objective evolutionary algorithm \cite{sarro2016multi}, or based on prior estimates \cite{jorgensen2003effort}. We note that linear regression is not typically used in the construction of prediction intervals with respect to software engineering predictions. In our case, our datasets violated multiple assumptions of linear regression such as normality and homoscedasticity; thus, we did not consider it as a viable approach.

\citet{thelin2002confidence} show that model-based point estimates, when used with confidence intervals, outperform human-based estimates. This result is particularly relevant as previous studies conclude that human-based estimates outperform model-based point estimates. 

\citet{vander1993assessing} evaluate Walds Likelihood CI and Mh–JK CIs and conclude that the Walds Likelihood CI includes the correct number of defects in most cases, and therefore should be preferred over Mh-JK. However, the Walds Likelihood CI is often too conservative, which leads to large intervals \cite{petersson2004capture}. 

Notably, \citet{angelis2000simulation} describe an instance-based learning approach used in combination with bootstrapping to construct effort intervals. They claim to construct CIs, however, our reading of the paper suggests they actually construct PIs. 

\citet{jorgensen2003effort} introduce an approach used to construct effort PI based on the assumption that they are able to select prior projects with similar estimation errors to that of a future project, i.e., the selected projects must have similar degrees of uncertainty as the future project.

Regarding PI construction via expert-opinion elicitation, \citet{jorgensen2002combination} use human-based PIs in effort estimation and conclude that group-discussion based PIs are more effective than “mechanically” combining individual PIs. However, the approach they put forth requires significant time and effort on the part of the experts involved and is sensitive to the accuracy of human estimates. Further, PIs based on human estimates tend to be too narrow to reflect high confidence levels (e.g., 90\% nominal confidence) \cite{jorgensen2004better, connolly1997decomposed}.

In conclusion, despite significant research advances in predictive models for software analytics, there have been very few studies that investigate the use of model-based intervals (CI or PI). A majority of those that have rely on capture-recapture methods \cite{petersson2004capture, thelin2002confidence, biffl2001evaluating, thelin2004team, padberg2002empirical}, which are a very specific type of prediction and have the distinct disadvantage of requiring human review. Methods relying on human estimates \cite{thelin2002confidence, jorgensen2002combination, briand1998cobra} are infeasible in our context. To our knowledge, no SE studies have investigated constructing PIs with Random Forest. 
\subsection{Random Forest}\label{sec:related_rf}
Decision trees are attractive for their execution time and the ease with which they can be interpreted. However, decision trees tend to overfit and generalize poorly to unseen data \cite{witten2016data}. The Random Forest (RF), first introduced in 1995 by Tin Kam Ho \cite{ho1995random}, is an example of ensemble learning in which multiple decision trees are grown independently. 

Given  data-points $D_i$ and responses $y_i$, $i=1, \dots, m$, each decision tree is grown from a bootstrapped sample of \textit{D}. Further, only a random subset of predictors is considered for the split of each node \cite{breiman2001random}. The number of predictors to consider at each split is specified by the parameter \textit{MTRY}. 

Thus, the trees grown in different, random subspaces generalize in a complementary manner, and their combined decision can be monotonically improved \cite{ho1995random, breiman2001random}. Given data-points $D_i$ and responses $y_i$, $i=1, \dots, m$, where each $D_i$ contains $n$ predictors, each tree in the RF is grown as follows:
\begin{itemize}
    \item Construct the training set by sampling, with replacement, \textit{m} data-points from \textit{D}.
    \item At each node, $n\textprime = n \times MTRY$ predictors are selected randomly and the most informative of these \textit{n\textprime} predictors is used to split the node.
    \item Each tree is grown to the largest extent possible.
\end{itemize}
The prediction $\hat{y}$ of a single tree $T$ for a new data-point $\vec{X}=\vec{x}$ is obtained by averaging the observed values $\ell(\vec{x})$, where $\ell$ is the leaf that is reached when dropping $\vec{x}$ down $T$. 

The prediction of a Random Forest $\hat{y}$ is equivalent to the conditional mean $\hat{\mu}(\vec{x})$ which is approximated by averaging the prediction of the $k$ trees \cite{breiman2001random}:
\begin{equation}\label{eq:mu}
    \hat{y} = \hat{\mu}(\vec{x}) = k^{-1}\sum_{i=1}^{k}\ell_k(\vec{x}).    
\end{equation}

\citet{guo2004robust} were the first to investigate the use of RF in defect classification. They conclude that RF outperforms the logistic regression and discriminant analysis of SAS\footnote{http://www.sas.com}, the VF1 and VotedPerceptron classifiers of WEKA\footnote{http://www.cs.waikato.ac.nz/ml/weka/}, and NASA's ROCKY \cite{menzies2003can}.
\subsection{Random Forest Prediction Intervals}\label{sec:related_qrf}

\begin{figure}[!t]
    \centering
    \includegraphics[width=.5\columnwidth]{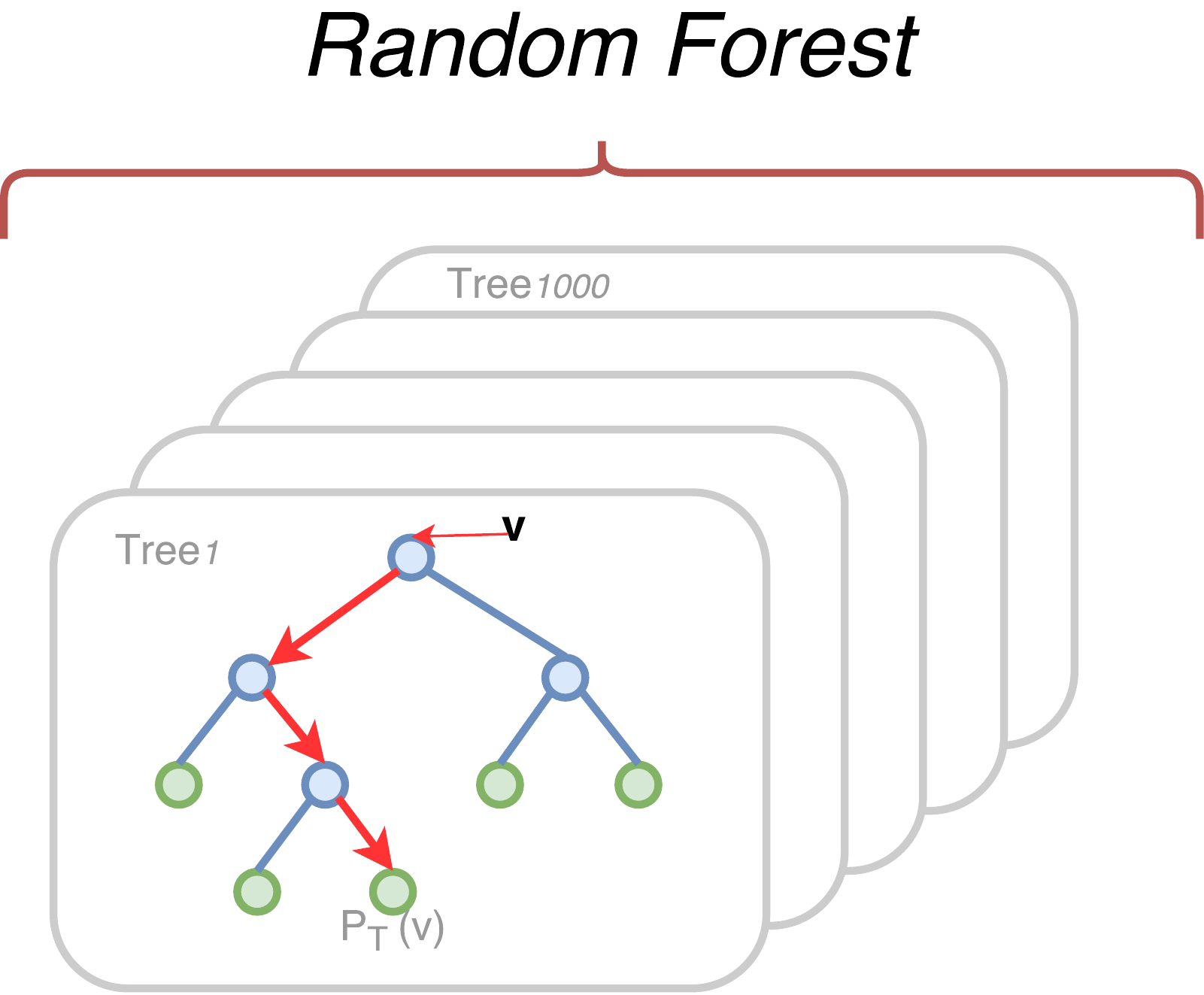}
    \caption{Random Forest overview.}
    \label{fig:randomforest}
\end{figure}
A common goal in statistical analysis is to infer the relationship between a response variable, $Y \in\mathbb{R}$, and a predictor variable $\vec{X}$. Given $\vec{X} = \vec{x}$, typical regression analysis determines an estimate $\hat{\mu}(\vec{x})$ of the conditional mean of $Y$. However, the conditional mean describes just one aspect of the relationship, neglecting other features such as fluctuations around the predicted mean \cite{meinshausen2006quantile}. In general, given $\vec{X} = \vec{x}$ the distribution function is defined by the probability that $Y \leq y$ \cite{evans2000statistical}:
\begin{equation}
    F(y|\vec{x}) = P(Y \leq y|\vec{X} = \vec{x}).
\end{equation}
For a continuous distribution, the $\tau$-quantile is defined such that $\tau = F(y|\vec{X} = \vec{x})$ \cite{meinshausen2006quantile}:
\begin{equation}
    Q_\tau(\vec{x}) = inf\{y: F(y|\vec{X} = \vec{x}) \geq \tau\}.
\end{equation}
We can use these definitions to construct PIs. For instance, the 90\% prediction interval is given by:
\begin{equation}
    I_{.90}(\vec{x}) = [Q_{.05}(\vec{X} = \vec{x}), Q_{.95}(\vec{X} = \vec{x})]
\end{equation}
In other words, there is a 90\% probability that a new observation of $Y$ is contained in the interval. \citet{meinshausen2006quantile} presents the Quantile Regression Forest (QRF), a general method for constructing prediction intervals based on decision trees. Rather than approximating the conditional mean, he shows that the responses from each tree $T_i$, $i=1,\dots,k$ can be used to approximate the full conditional distribution function $\hat{F}(y|\vec{X} = \vec{x})$. 

We can use the RF to construct prediction intervals if we ensure that all trees are fully expanded, i.e., each leaf has only one response. For a given $\vec{X} = \vec{x}$ and RF with trees $T_i$, $i=1, \dots, k$, the distribution function can be approximated as:
\begin{equation}
    \hat{F}(y|\vec{x}) = k^{-1}\sum_{i=1}^{k}1_{\{\ell_i(\vec{x}) \leq y\}}.
\end{equation}
In other words, the likelihood that the prediction $\hat{y}_i = \ell_i(\vec{x})$ is less than or equal to the response $y$. Then, the $\tau$-quantile can be approximated as:
\begin{equation}
    \hat{Q}_\tau(\vec{x}) = inf\{y: F(y|\vec{X} = \vec{x}) \geq \tau\}.
\end{equation}
Finally, the 90\% prediction interval can be approximated as:
\begin{align}\label{equation:intervals}
    \hat{I}_{.90}(\vec{x}) = [\hat{Q}_{.05}(\vec{X} = \vec{x}), \hat{Q}_{.95}(\vec{X} = \vec{x})].
\end{align}
We note that it is possible that we are unable to fully expand all of the trees. This can happen if (1) the node is already pure, or (2) the node cannot be split further, i.e., all remaining data-points have the same predictors. In the former case, we know the response and the leaf size, i.e., we can still determine the conditional distribution. In the latter case, we do not know what the response would be if the node is split further; hence, we are unable to determine the conditional distribution. Lastly, we note that fully-expanding trees can lead to overfitting, in which case the intervals will provide little practical value \cite{meinshausen2006quantile}. However, fully expanding trees is what \citet{breiman2001random} suggests. Further, even if individual trees overfit, the effect should be mitigated by increasing the total number of trees in the forest.

\begin{algorithm}[!h]
\caption{Random Forest Prediction Intervals}\label{alg:qrf}
\begin{algorithmic}[1]
\Procedure{PredictionIntervals}{$D$, $y$, $nc$, $ph$}
    \State $\textit{D\_train}, \textit{D\_test}, \textit{y\_train}, \textit{y\_test} \gets \text{split(D, y, ph)}$\label{line:split}
    \State $\textit{clf} \gets \text{RandomForestRegressor()}$
    \State $\textit{clf.fit(D\_train, y\_train)}$\label{line:fit}
    \\
    \State $preds \gets [[]]$
    \ForAll{$\textit{est}~\textbf{in}~\text{clf.estimators}$}\label{line:start}
        \ForAll{$\textit{n, leaf}~\textbf{in}~\text{est.apply(D\_test)}$}
            \If{\textit{leaf}~\textbf{is not pure}}
                \State \textbf{exit} \label{line:check_pure}
            \EndIf
            \State $\textit{preds[n]}.extend(\textit{leaf.responses()})$
        \EndFor
    \EndFor\label{line:end}
    \\
    \State $\alpha \gets \frac{1 - nc}{2}$
    \State $\beta \gets 1 - \alpha$
    \State $intervals \gets []$
    \ForAll{$\textit{pred}~\textbf{in}~\text{preds}$}
        \State $intervals.append((Q_{\alpha}(pred), Q_{\beta}(pred)))$\label{line:intervals}
    \EndFor
    \State $\textbf{return}~\textit{intervals}$
\EndProcedure
\\
\Procedure{}{}
\EndProcedure
\end{algorithmic}
\end{algorithm}
We present the procedure for constructing prediction intervals in Algorithm~\ref{alg:qrf}. The procedure accepts four arguments: \textit{D} (the data-points),  \textit{y} (the responses), \textit{nc} (the nominal confidence), and \textit{ph} (the percent to be held out). In lines \ref{line:split}-\ref{line:fit} we partition the data into training and test sets and learn a RF. In lines \ref{line:start}-\ref{line:end} we record the responses of all decision trees for each $d_i \in D_{test}$. We ensure that each leaf is pure on line~\ref{line:check_pure}. Finally, we construct prediction intervals on line~\ref{line:intervals}.

Table~\ref{tbl:sklearn_rf_params} presents the list of parameters for Sci-Kit Learn's RandomForestRegressor.

\begin{table}[!t]
\centering
\caption{Sci-Kit Learn RandomForestRegressor parameters}
\label{tbl:sklearn_rf_params}
\resizebox{.7\columnwidth}{!}{%
\begin{tabular}{|p{.2\linewidth}|p{.4\linewidth}|c|}
\hline
feature & description & value \\ \hline
\textbf{n\_estimators} & \textbf{The number of trees in the forest.} & \textbf{1000} \\ \hline
\textbf{max\_features} & \textbf{The size of the subset of the features to consider for each split.} & \textbf{1.0*} \\ \hline
max\_depth & The maximum depth of the tree. & None \\ \hline
\makecell{min\_samples\_\\split} & The minimum number of samples required to split an internal node. & 2 \\ \hline
\makecell{min\_samples\_\\leaf} & The minimum number of samples required to be at a leaf node. & 1 \\ \hline
\makecell{min\_weight\_\\fraction\_leaf} & The minimum weighted fraction of the sum total of weights (of all the input samples) required to be at a leaf node. & 0 \\ \hline
max\_leaf\_nodes & If specified, trees are grown with the specified number of leaves in a best first fashion. & None \\ \hline
\makecell{min\_impurity\_\\split} & Threshold for stopping early tree growth. & 1E-7 \\ \hline
\end{tabular}}
\end{table}


\subsection{Validation Techniques}\label{sec:related_validation}
Validation techniques (also referred to as performance estimation techniques) are methods used for estimating performance of a model on unseen data. They are widely used in the machine learning domain to compare the performance of multiple models on a dataset in order to select the most well suited model (i.e., the model that minimizes the expected error) \cite{d2012evaluating, lessmann2008benchmarking, myrtveit2005reliability}.

Within the context of validating defect prediction models, \citet{tanti2017empirical} report 89 studies (49\%) use k-fold cross-validation, 83 studies (45\%) use holdout validation, 10 studies (5\%) use leave-one-out cross-validation, and 1 study (0.5\%) uses bootstrap validation.

Validation techniques are also used in hyper-parameter optimization, i.e., tuning the parameters of a model \cite{tantithamthavorn2016automated, fu2016tuning}. There are many validation techniques that can be applied and no consensus on which technique is the most effective \cite{beleites2005variance, braga2004cross, breiman1992submodel, kocaguneli2013software, kohavi1995study, mockus2000predicting, tanti2017empirical}. 

To our knowledge, only two techniques have been used in tuning models for software analytics: \citet{fu2016tuning} use the holdout and \citet{tantithamthavorn2016automated} use bootstrap.  These studies serve as inspiration for this work; we analyze the two techniques used in their studies and include an additional six more. Moreover, we reuse part of their experimental design: we holdout one third of the initial dataset for testing. One of the major differences in this paper is that both our datasets contain data-points measured across multiple industrial projects and multiple releases of the same project, whereas each of their datasets contain data-points measured from a single release of an open-source project. Thus, our datasets are sensitive to time and this aspect could affect the performance of certain tuning techniques that do not preserve the dataset order (e.g., bootstrap) in learning and testing.

\subsubsection{Order Preserving}
Techniques which preserve the order of the dataset are commonly used for performance estimation with time-series data (e.g., forecasting) \cite{bergmeir2012use}. We analyze five such techniques: \textit{time series cross-validation}, \textit{time series HV cross-validation}, and three variations of the \textit{non-repeated holdout}. Given data-points $D_i$ and responses $y_i$, $i=1, \dots, m$:

\description{H/k holdout} partitions the dataset such that $\frac{h}{m}$ data-points are used for training and $\frac{k}{m}$ are used for testing. Holdout validation has the advantage of being faster than techniques which require multiple repetitions and is often acceptable to use if $D$ is sufficiently large \cite{witten2016data}. However, it is criticized for providing unreliable estimates \cite{tanti2017empirical} due to the fact that performance is estimated on a single subset of $D$. Further, the holdout is criticized as being statistically inefficient since much of the data is not used to train the model \cite{tanti2017empirical}. Figure~\ref{fig:data} provides an example of 66/33 holdout between validation-training and validation-testing.

\description{Time series cross-validation} (TSCV) computes an average of k-step ahead forecasts by the rolling-origin-calibration procedure \cite{borra2010measuring}. Note that unlike standard cross-validation procedures, the $i+1$ training set is a super-set of all training sets $[1, 2, \dots, i]$. TSCV has the advantage of being relatively fast. However, theoretical concerns have been raised about dependencies between the data-points in the training and test sets \cite{bergmeir2012use}. 

\description{Time series HV cross-validation} \cite{racine2000consistent} (TSHVCV) is an extension of the time series h-block cross validation introduced by \citet{burman1994cross} in \citeyear{burman1994cross}. TSHVCV removes $v$ data-points from both sides of $D_i$, yielding a validation set of size $2v + 1$. Then, $h$ data-points on either side of the validation set are removed. The remaining $n - 2v - 2h - 1$ data-points are used in training. The process is repeated for $i \in [v+0s, v+1s, \dots, m-v]$ where $s$ is a specified step size. TSHVCV has the benefit of being theoretically robust to data dependencies. However, it can be computationally expensive if $D$ is sufficiently large.   

\subsubsection{Non-Order Preserving}
Techniques which do not preserve the order of the dataset are commonly used in the general context of machine learning validation and are particularly effective when data-points are not time sensitive. An example of non-sensitive data is the characteristics of classes and their defectiveness within a single software release of an open-source project. We analyze three such techniques: $h \times k$\textit{-fold cross-validation}, \textit{K out-of-sample bootstrapping}, and \textit{leave-one-out cross-validation}. Given data-points $D_i$ and responses $y_i$, $i=1, \dots, m$:

\description{K-fold cross-validation} partitions the data into $k$ equal sized \textit{folds}: $k-1$ folds are used for training and the remaining fold is used for testing. This is repeated until each fold has been used for testing. $H \times k$-fold cross-validation performs k-fold cross-validation $h$ times, re-seeding the PRNG at each iteration. Cross-validation techniques have the advantage of utilizing all of the data-points in $D$ for training. However, if $D$ is time sensitive, cross-validation will construct training sets containing data-points measured after data-points in the test set. Further, cross-validation techniques have been shown to produce unreliable estimates if $D$ is small \cite{jiang2009variance, braga2004cross}. 

\description{K out-of-sample bootstrapping} is a technique in which the dataset is randomly sampled with replacement. The training set is constructed by randomly sampling $m$ data points with replacement from $D$. The out of bag samples are used for testing. The bootstrap has the advantage of being more effective at providing accurate estimates when $D$ is small. However, the bootstrap can be misleadingly optimistic \cite{witten2016data}.

\description{Leave-one-out cross-validation} (LOO) is a case of k-fold cross-validation in which $k=1$. LOOCV has the advantage of using the greatest possible amount of data-points in training at each iteration. Further, prior work in effort estimation has shown that LOOCV is the least biased technique \cite{kocaguneli2013software}. However, LOOCV has also been criticized for providing estimates with high variance \cite{friedman2001elements}.
\vspace{-.5\baselineskip}
\subsection{Tuning}\label{sec:related_tuning}
Recent software engineering studies indicate that configuring models can impact model performance \cite{koru2005investigation, mende2010replication, fu2016tuning, tosun2009reducing, tantithamthavorn2016automated}. 

\citet{fu2016tuning} investigate whether tuning is necessary in software analytics. They find that tuning can increase model performance by 5-20\%, and in one case they show Precision can be improved from 0\% to 60\%. They report that 80\% of highly-cited defect prediction studies use default model configurations and raise concerns about conclusions drawn from these studies regarding the superiority of one model over another. They conclude that tuning is a necessary step before applying a model to the task of defect prediction.

\citet{tosun2009reducing} investigate tuning with respect to defect classification and found that using the default model configuration can produce sub-optimal results. Specifically, they find that tuning the decision threshold of the Naive Bayes Classifier can decrease false alarms by 34\% to 24\%.

\citet{tantithamthavorn2016automated} show that configuring models can increase performance by as much as 40\% and increases the likelihood of producing a top-performing model by as much as 83\%. Similar to \citet{fu2016tuning}, they also conclude that tuning is a necessary step in defect prediction studies.  

Thus, as far as we know,
no study has investigated the tuning of RF prediction intervals. Moreover, to our knowledge, no study has compared the performance of multiple tuning techniques in the context of software engineering. Finally, no study has investigated meta-tuning in the context of software engineering.

\section{Experimental Design}\label{sec:design}
\subsection{Objects of Study}
In this section, we describe the characteristics of the two datasets we consider in this study.
\subsubsection{Keymind}\label{sec:design_keymind}
The industry context motivating this study is Keymind, a CMMI Level 5 organization \cite{chrissis2011cmmi}. Keymind is the technology and creative services division of Axiom Resource Management, a professional consulting firm based in Falls Church, Virginia. Keymind provides software development, strategic marketing, and user focused services to both commercial and government customers. The Keymind staff comprises award-winning leaders in interactive media who are widely recognized for distinctive website designs. Keymind focuses on a user-friendly experience, accessibility, and client-focused creative and technical support. It is a narrow to medium size organization of about 25-30 people and was successfully re-appraised at Maturity Level 5 of the Capability Maturity Model Integration for Development (CMMI-DEV), version 1.3 in September, 2015. Keymind relies heavily on data collected from its suite of development environment tools to support quantitative project management and organizational decision-making. Their development environment tool suite includes: 1) JIRA \cite{doar2011practical}, a web-based issue-tracking tool; 2) Subversion (SVN) \cite{pilato2008version}, integrated with JIRA, for configuration management including source code and documentation control; 3) Confluence \cite{kohler2013atlassian}, a wiki that serves as a process asset library and knowledge sharing repository; 4) Jenkins \cite{smart2011jenkins} for supporting a continuous integration and unit testing automation; and 5) SonarQube \cite{campbell2013sonarqube}  for managing technical debt.

The Keymind dataset consists of 363 data-points (i.e., project releases) collected from 13 projects spanning the course of 13 years (i.e., 2003 to 2016). Each data-point consists of 12 predictors. We worked closely with the Keymind team in the construction of the dataset. In 2016, about 10 Keymind employees, comprising project managers and developers, performed a survey in which they identified factors that they believe influence the number of post-release defects. 

We reviewed the results of the survey and filtered out predictors that were unknowable or inactionable at the time of release definition. Regarding being knowable, a well-known metric impacting the defect proneness of a system is the number of god classes \cite{lanza2007object, falessi2015validating}. However the user only knows the number of god classes at the beginning of the release, not at its end. Therefore, we included as a predictor the number of god classes at the beginning of a release, not at its end. Regarding being actionable, a well-known metric impacting the defect proneness of a system is the number of developers \cite{lanza2007object}. Thus, during release definition, the user wants to specify the number of developers to assign to the release. Finally, it is well known that size impacts defect proneness; however, the typically used size-metric, lines of code (LOC), has two problems: (1) it is unknown during release definition, and (2) it is not reliable when different projects make different use of automatically generated code. Thus, rather than using LOC as metric of size, we use the number of change requests and number of implementation tasks. 

It was also important that the tool is easy to use, and users suggested that there should be no more than 15 predictors. Eventually we selected a total of 12 predictors. Table~\ref{tbl:keymind_data} presents summary statistics for each of the selected predictors. The appendix reports the definition and the rationale of each predictor.

We looked for outliers and we eventually decided to not remove any data points from the analysis. The rationale is that, after a careful analysis, all data points were determined to be correct, i.e., they represent an event that actually happened and is reasonable to predict. This is especially true given that Keymind is CMMI Level 5 certified and it significantly relies on data analysis during software development.

\begin{table}[!b]
\centering
\caption{Keymind dataset characteristics (after pre-processing)}
\label{tbl:keymind_data}
\begin{tabular}{|l|l|c|c|c|c|}
\hline
Predictor & Type & Min & Mean & Max & $\rho$ \\ \hline
\begin{minipage}{.2\columnwidth}Project Age\end{minipage} & numeric & -204.0 & 1110.93 & 4456.0 & -0.13* \\ \hline
\begin{minipage}{.2\columnwidth}Development Duration\end{minipage} & numeric & -2.97 & 1.04 & 36.73 & 0.18* \\ \hline
\begin{minipage}{.2\columnwidth}Estimated Effort\end{minipage} & numeric & 0.0 & 32.26 & 520.5 & 0.41* \\ \hline
\begin{minipage}{.2\columnwidth}Testing Time\end{minipage} & numeric & 0.0 & 0.37 & 3.79 & -0.13* \\ \hline
\begin{minipage}{.2\columnwidth}Number of Change Requests\end{minipage} & numeric & 0.0 & 6.35 & 65.0 & 0.42* \\ \hline
\begin{minipage}{.2\columnwidth}Number of Developers\end{minipage} & numeric & 0.0 & 115.49 & 752.72 & 0.17* \\ \hline
\begin{minipage}{.2\columnwidth}Number of Implementation Tasks\end{minipage} & numeric & 0.0 & 3.41 & 79.0 & 0.27* \\ \hline
\begin{minipage}{.2\columnwidth}Number of New Developers\end{minipage} & numeric & 0.0 & 0.28 & 7.0 & 0.14* \\ \hline
\begin{minipage}{.2\columnwidth}Number of Release Candidates\end{minipage} & numeric & 0.01 & 0.07 & 1.0 & 0.03 \\ \hline
\begin{minipage}{.2\columnwidth}Number of Users\end{minipage} & numeric & 3.0 & 149.95 & 700.0 & 0.21* \\ \hline
\begin{minipage}{.2\columnwidth}Prior Number of God Classes\end{minipage} & numeric & 0.0 & 21.98 & 57.0 & 0.30* \\ \hline
\begin{minipage}{.2\columnwidth}Project Architecture\end{minipage} & \makecell{categ-\\orical} &  &  &  &  \\ \hline
\end{tabular}
\end{table}

\subsubsection{Tukutuku}
In addition to the Keymind dataset, we use the Tukutuku dataset as provided to us by the first author of \citet{mendes2005investigating}. The Tukutuku Research Project aims to gather data from Web companies to benchmark development productivity and develop cost estimation models.   

The dataset consists of 195 data-points collected from 67 industrial Web projects. Each data-point consists of 15 predictors. In general, the predictors are categorized as static (e.g., total number of web pages, total number of images), dynamic (e.g., types of features/functionality), or project (e.g., size of development team, development team experience). We refer the reader to \citet{mendes2005investigating} for a more detailed description of the dataset and context. 
\subsection{RQ1: Is default the best configuration?}\label{sec:rq1_1_design}

In the machine learning domain, the importance of tuning is well known \cite{bergstra2011algorithms, snoek2012practical, duan2003evaluation}. Moreover, several recent SE studies strongly suggest that model performance can be improved by tuning \cite{fu2016tuning, tantithamthavorn2016automated, borg2016tuner}. With respect to RF, \citet{meinshausen2006quantile} argues that results are typically near optimal over a wide range of MTRY, suggesting that all configurations are the same. This is supported by \citet{breiman1992submodel} who show that RF performance is insensitive to the number of predictors considered at each split. On the other hand, \citet{guo2004robust} claim that the optimal MTRY is $\sqrt{n}$ where $n$ is the number of predictors.

If default is the best configuration, then there is no need for tuning and/or configuration validation techniques. If default is not the best configuration, then it makes sense to investigate how to effectively and automatically identify the best configuration (discussed further in subsequent research questions). Validation techniques could also be useful for identifying cases where all configurations are unreliable, hence warning that we cannot be confident about PI coverage. Thus, in this research question, we investigate the coverage and width of the different configurations to determine if tuning RF is promising, i.e., if there are cases where default is not the best configuration or all configurations are unreliable. 

\subsubsection{Hypotheses}\label{sec:rq1_1_hypotheses}
We conjecture that default is not the best configuration among all NCs and datasets and that the coverage and width of PIs differ among configurations.

\begin{nullhyp}\label{nh:rq1_1_coverage}
The coverage is equal among configurations.
\end{nullhyp}
\begin{nullhyp}\label{nh:rq1_1_width}
The width is equal among configurations.
\end{nullhyp}
\subsubsection{Independent Variables}\label{sec:rq1_1_independent}
The independent variable in this research question is the RF configuration. Among the different RF parameters (see Table~\ref{tbl:sklearn_rf_params}), in this work, we only vary \textit{max\_features} (MTRY) and we set \textit{n\_estimators} to 1000 as \citet{meinshausen2006quantile} suggests sufficiently large forests are necessary for good intervals. MTRY specifies the size of the subset of predictors that will be considered as candidates for splitting each node. The default MTRY for Sci-Kit Learn's RandomForestRegressor is 1.0, i.e., 100\% of predictors are considered at each split.  We recognize it is theoretically possible that configuring parameters other than MTRY could affect the benefit of validation or meta-validation techniques. However, we do not configure parameters other than MTRY because there is evidence that this would increase the likelihood that the leaves of the trees would be impure, and hence it would not be possible to infer valid quantile distributions (see Section~\ref{sec:related_qrf}). Thus, the treatments are 20 values of MTRY (i.e., [0.05, 1.0] step 0.05). We select a step of 0.05 on MTRY to ensure that all integer numbers of predictors are analyzed, i.e., an effectively exhaustive search. We note that the search need only enumerate a small grid (i.e., with $n$ predictors: $\{i \mid i \in \mathbb{Z}^+, i \leq n\}$), and therefore does not need to be directed.

\subsubsection{Dependent Variables}\label{sec:rq1_1_dependent}
The dependent variables are Coverage, Width and Potential Benefit. 

Given data-points $D_i$, responses $y_i$, $i=1, \dots, m$, and a nominal confidence $\varphi$, we define point-coverage $C_i$ as a boolean value for each data-point:
\begin{equation}
    C_i = 
    \begin{cases}
    1,& \text{if }y_i \in I_\varphi(D_i)\\
    0,& \text{otherwise}
\end{cases}.
\end{equation}

We define coverage $C$ as the proportion of times the true values are contained in the intervals:
\begin{equation}
    C = m^{-1}\sum_{i=1}^{m}C_i.
\end{equation}

Let $\alpha = \frac{1 - \varphi}{2}$ and $\beta = 1 - \alpha$ be the quantiles of the lower and upper bounds of $I_\varphi$, respectively. We define point-width $W_i$ as the $\beta_i$-quantile minus the $\alpha_i$-quantile (see Section~\ref{sec:related_qrf}):
\begin{equation}
    W_i = Q_\beta(D_i) - Q_\alpha(D_i).
\end{equation}   
We define mean width $MW$ as the average of all point-widths:
\begin{equation}
    MW = m^{-1}\sum_{i=1}^{m}W_i.
\end{equation}

The objective function (i.e., smallest reliable width) is a factor of two metrics: coverage and width. Therefore, to determine if default is the best configuration, we define the following tags, in descending order of potential benefit, according to the possible scenarios of the performance of the default configuration relative to other configurations.
\begin{itemize}
    \item \textbf{DU}: the \textit{default} configuration is \textit{unreliable} and there exists at least one other reliable configuration. This scenario has the most potential benefit as PIs are unreliable without tuning.
    \item \textbf{SB}: the default configuration is reliable and there exists some other reliable configuration with \textit{better} (i.e., smaller) mean width and the \textit{difference is significant}. This scenario has less potential benefit than DU as PIs are still reliable without tuning, however the intervals constructed are not as narrow as possible.
    \item \textbf{AU}: \textit{all} configurations are \textit{unreliable}. This scenario has less potential benefit than SB as PI are unreliable, however tuning can still prove beneficial if we recognize this case, i.e., we can warn the user that we cannot support a sentence like ``We are $\varphi$\% confident that the number of defects is between $a$ and $b$.''
    \item \textbf{NSB}: the default configuration is reliable and there exists some other reliable configuration with \textit{better} mean width but the \textit{difference is not significant}. This scenario has no potential benefit because, practically speaking, default is equivalent to the best configuration.
    \item \textbf{E}: the default configuration is reliable and has the smallest mean width. This scenario has no potential benefit.
\end{itemize}
To summarize, \textit{DU}, \textit{SB}, and \textit{AU} are scenarios in which the default configuration is considered sub-optimal, whereas \textit{NSB} and \textit{E} are scenarios in which the default configuration is considered optimal.

\subsubsection{Analysis Procedure}\label{sec:rq1_1_procedure}
Given data-points $D_i$, responses $y_i$, $i=1, \dots, m$, and $MTRY = [0.05, 0.10, \dots, 1.0]$, we apply the following procedure for each $mtry$ in $MTRY$.
\begin{enumerate}
    \item Perform a 66/33 holdout to partition $D$ into \textit{train} and \textit{test} (see Figure~\ref{fig:data}). 
    \item Learn a RF on \textit{train} and evaluate Coverage and mean width (MW) on \textit{test}.
    \item We report the coverage and mean width distributions. Additionally, we analyze coverage and mean width of configurations and report the scenario tag.  
\end{enumerate}
We test \textit{H\ref{nh:rq1_1_coverage}0} by applying Cochran's Q test on the point-coverage of all configurations. We test \textit{H\ref{nh:rq1_1_width}0} by applying Friedman's test on the point-widths of all configurations. Cochran's Q and Friedman's are paired, non-parametric statistical tests, and hence fit our experimental design. We use an alpha of 0.05 in these and all subsequent statistical tests, i.e., we reject a null hypothesis if $p\leq 0.05$. We perform all analysis separately for each NC and dataset.
\subsection{RQ2: Are validation techniques beneficial?}\label{sec:rq1_2_design}
In Section~\ref{sec:rq1_1_design}, even if default is not the best configuration, it could be the case that using a validation technique is counterproductive for tuning. For example, the validation technique might select a configuration that is unreliable or with larger mean width than the default configuration. Thus, in this research question, we investigate the benefit provided by eight different validation techniques.  

\subsubsection{Independent Variables}
The independent variable in this research question is the validation technique used for tuning.

The treatments are the eight validation techniques presented in Section~\ref{sec:related_validation} in combination with the following selection heuristics.
\begin{itemize}
    \item If one or more configurations are predicted to be reliable, select the configuration with the smallest predicted width.
    \item If all configurations are predicted to be unreliable, indicate that no configuration should be used.
\end{itemize}

\subsubsection{Dependent Variables}
The dependent variable in this research question is Level of Benefit. Similar to RQ~\ref{rq:tune_promise}, we define tags according to the possible scenarios of the default performance relative to the configuration selected by a technique.
\begin{itemize}
    \item \textbf{DU}: the \textit{default} configuration is \textit{unreliable} and the configuration selected by the technique is reliable. This scenario is by far the most beneficial as PI are unreliable without the use of a validation technique.
    \item \textbf{SB}: the default configuration is reliable and the selected configuration is reliable and has \textit{better} (i.e., smaller) mean width and the \textit{difference is significant}. This scenario is less beneficial than DU since the use of a validation technique is not necessary to produce reliable PI, however tuning is still beneficial since we can construct narrower PI.
    \item \textbf{APU}: all configurations, including default, are \textit{actually and predicted to be unreliable}. This scenario is less beneficial than SB as PI are unreliable, however tuning is still beneficial since we know that PI are unreliable.
    \item \textbf{PU}: all configurations, including default, are \textit{predicted to be unreliable}, default is unreliable and there is at least one  configuration actually reliable. Despite the fact that the technique fails to identify any reliable configuration, this scenario is still beneficial as it correctly avoids the use of an unreliable default.
    \item \textbf{NSD}: the default configuration and the selected configuration are both reliable and are \textit{not significantly different} with respect to point-width. This scenario is neutral, i.e., tuning does not have any effect.
    \item \textbf{NKU}: the default configuration and the selected configuration are \textit{not known} to be \textit{unreliable}. This scenario is the least counterproductive since the use of a validation technique led to unreliable PI, but the result would have been the same without the use of a validation technique.
    \item \textbf{SW}: the default configuration is reliable and there exists some other reliable configuration with \textit{worse} (i.e., larger) mean width and the \textit{difference is significant}. This scenario is more counterproductive than NKU since PI are worse as a result of using a validation technique.
    \item \textbf{TU}: the default configuration is reliable and the selected configuration is unreliable or no configuration was selected (not even default), i.e., the \textit{technique} is \textit{unreliable}. This scenario is the most counterproductive as PI becomes unreliable or unusable as a result of using a validation technique. 
\end{itemize}
To summarize, the use of a technique is beneficial in scenarios \textit{DU}, \textit{SB}, \textit{APU} and \textit{PU}; neutral in scenario \textit{NSD}; and counterproductive in scenarios \textit{NKU}, \textit{SW}, and \textit{TU}. 

\subsubsection{Analysis Procedure}\label{proc:tune}
Given data-points $D_i$, responses $y_i$, $i=1, \dots, m$, and techniques $T = [t_1, t_2, ... t_8]$, we apply the following procedure for each $t$ in $T$.
\begin{enumerate}
    \item Perform a 66/33 holdout to partition $D$ into \textit{train} and \textit{test}.\label{proc:tune1}
    \item Apply $t$ to \textit{train}. For example, if $t$ is \textit{50/50 holdout}, we construct \textit{train$^\prime$} and \textit{test$^\prime$} by performing a 50/50 holdout on $train$.\label{proc:tune2}
    \item For each NC: 
    \begin{enumerate}
        \item For each $mtry$, determine predicted coverage and mean width by learning a RF on \textit{train$^\prime$} and evaluating on \textit{test$^\prime$}.\label{proc:tune4}
        \item For each $mtry$, determine actual coverage and actual mean width by learning a RF on \textit{train} and evaluating on \textit{test}.\label{proc:tune5} 
        \item Apply the selection approach and determine scenario tag according to the definitions above.\label{proc:tune6}
    \end{enumerate}
    \item We report the scenario tags for each NC and dataset. 
\end{enumerate}
We perform Friedman's test on configuration point-widths to discriminate scenarios \textit{SB}, \textit{NSD}, and \textit{SW}. We use Friedman's test because it is a paired, non-parametric statistical test, and hence fits our experimental design.

\subsection{RQ3: Are meta-validation techniques  beneficial?}
Since there is more than one validation technique that we can use for tuning, in this research question we investigate whether or not we are able to automatically choose the most beneficial validation technique.

\subsubsection{Independent Variables}
The independent variable of this research question is the meta-validation technique. The treatments are three meta h/k-holdouts (i.e., 25/75, 50/50, and 75/25). Meta-tuning is computationally expensive, specifically with larger datasets. We consider 25/75 as it significantly reduces the amount of data that needs to be processed, and could still perform reasonably well with large datasets. 

\subsubsection{Dependent Variables}
The dependent variable in this research question is Level of Benefit as defined in Section~\ref{sec:rq1_2_design}. 

\subsubsection{Analysis Procedure}
Given data-points $D_i$, responses $y_i$, $i=1, \dots, m$, techniques $T = [t_1, t_2, ... t_8]$, and meta-techniques $M = [m_1, m_2, m_3]$, we apply the following procedure for each $m$ in $M$.
\begin{enumerate}
    \item Perform a 66/33 holdout to partition $D$ into \textit{train} and \textit{test}.
    \item Apply $m$ to $1$. For example, if $m$ is \textit{50/50 holdout}, we construct \textit{train$^\prime$} and \textit{test$^\prime$} by performing a 50/50 holdout on \textit{train}.
    \item For each $t$ in $T$, determine predicted technique benefit by performing Steps~\ref{proc:tune2}-\ref{proc:tune6} of Procedure~\ref{proc:tune} with the newly partitioned data.
    \item Construct a technique, $t\textprime$, by choosing, for each NC, the technique that is predicted to be the most beneficial. 
    \item Identify the tag for $T = [t\textprime]$ to determine actual benefit.  
\end{enumerate}
We perform the above analysis for each dataset separately. We report the scenario tags for each of the three meta-techniques in each NC and dataset. 

\section{Results and Discussion}\label{sec:results}
In this section, we present results and discussion for each research question.
\subsection{RQ1: Is default the best configuration?}\label{sec:rq1_1_results}
\subsubsection{Results}
Figure~\ref{fig:coverage} and Figure~\ref{fig:width} present distributions of coverage and the mean width, respectively, of the 20 configurations, at each NC, in Keymind and Tukutuku. We display the nominal confidence along the x-axis and the dependent variable (i.e., coverage and mean width) along the y-axis. In Figure~\ref{fig:coverage}, we indicate the percent of reliable configurations, for each NC and dataset, in parentheses along the x-axis. We test \textit{H\ref{nh:rq1_1_coverage}0} using Cochran's Q test, and \textit{H\ref{nh:rq1_1_width}0} using Friedman's test for each NC. We denote a p-value lower than alpha with an asterisk (*) along the x-axis. For instance, looking at Figure~\ref{fig:coverage} (Tukutuku) we can observe at NC95 that 30\% of configurations are reliable and that there is a significant difference in coverage among configurations. 

Table~\ref{tbl:tuning_potential_benefit} presents the tags related to the potential benefit (see Section~\ref{sec:rq1_2_design}) of tuning. Column 1 reports the dataset and columns 2, 3, and 4 report the tags at NC90, NC95, and NC99 respectively. To improve readability, we color coded the tags: the darker the color, the highest the potential benefit of tuning. For instance, looking at column 3 row 3 of Table~\ref{tbl:tuning_potential_benefit} we can observe that at NC95 in Tukutuku the default configuration is unreliable (and there exists at least one reliable configuration).

\subsubsection{Discussion}
\begin{figure}[!t]%
    \centering
    \subfloat[Keymind]{{\includegraphics[width=.7\columnwidth]{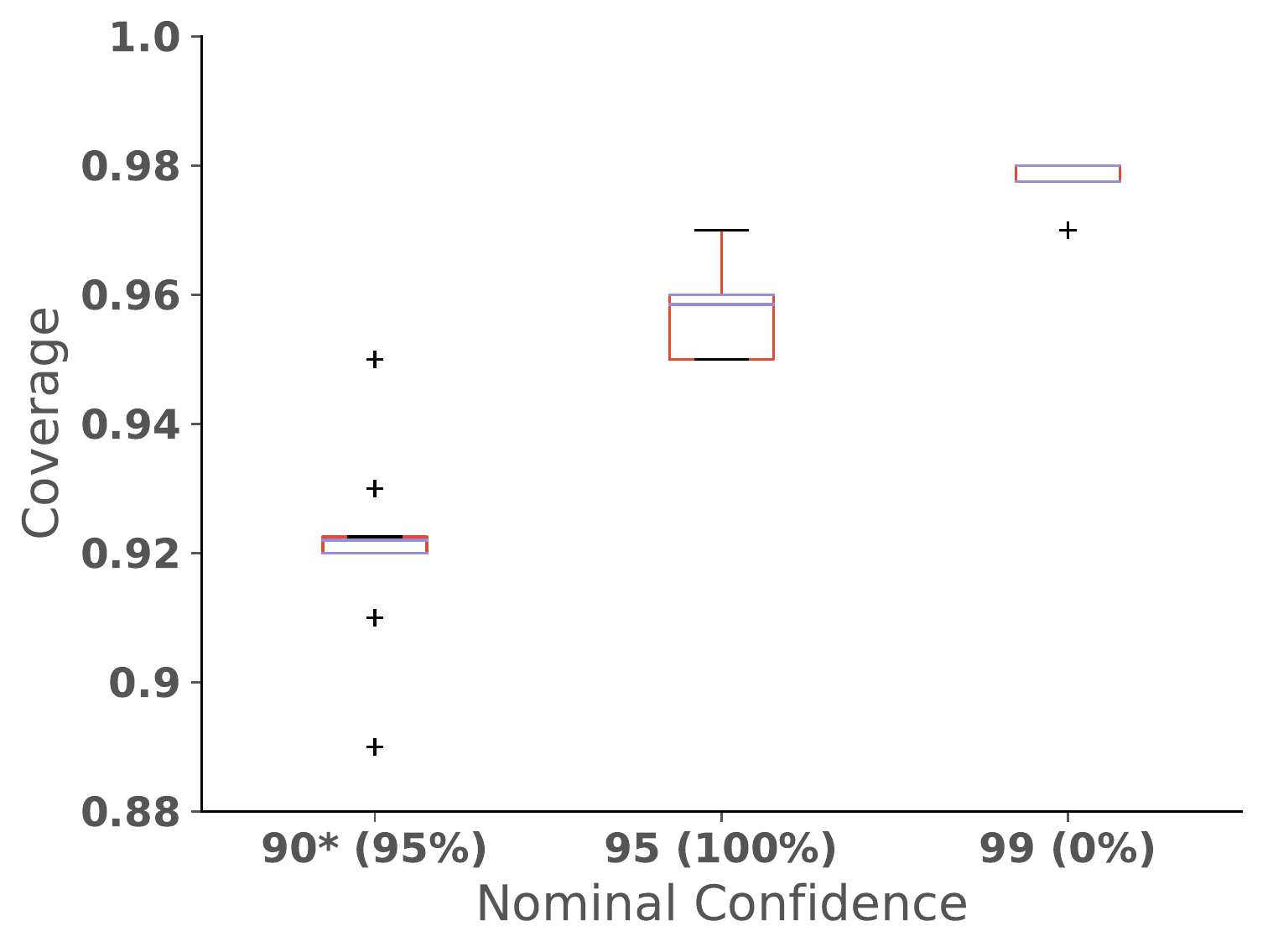} }}
    \\
    \subfloat[Tukutuku]{{\includegraphics[width=.7\columnwidth]{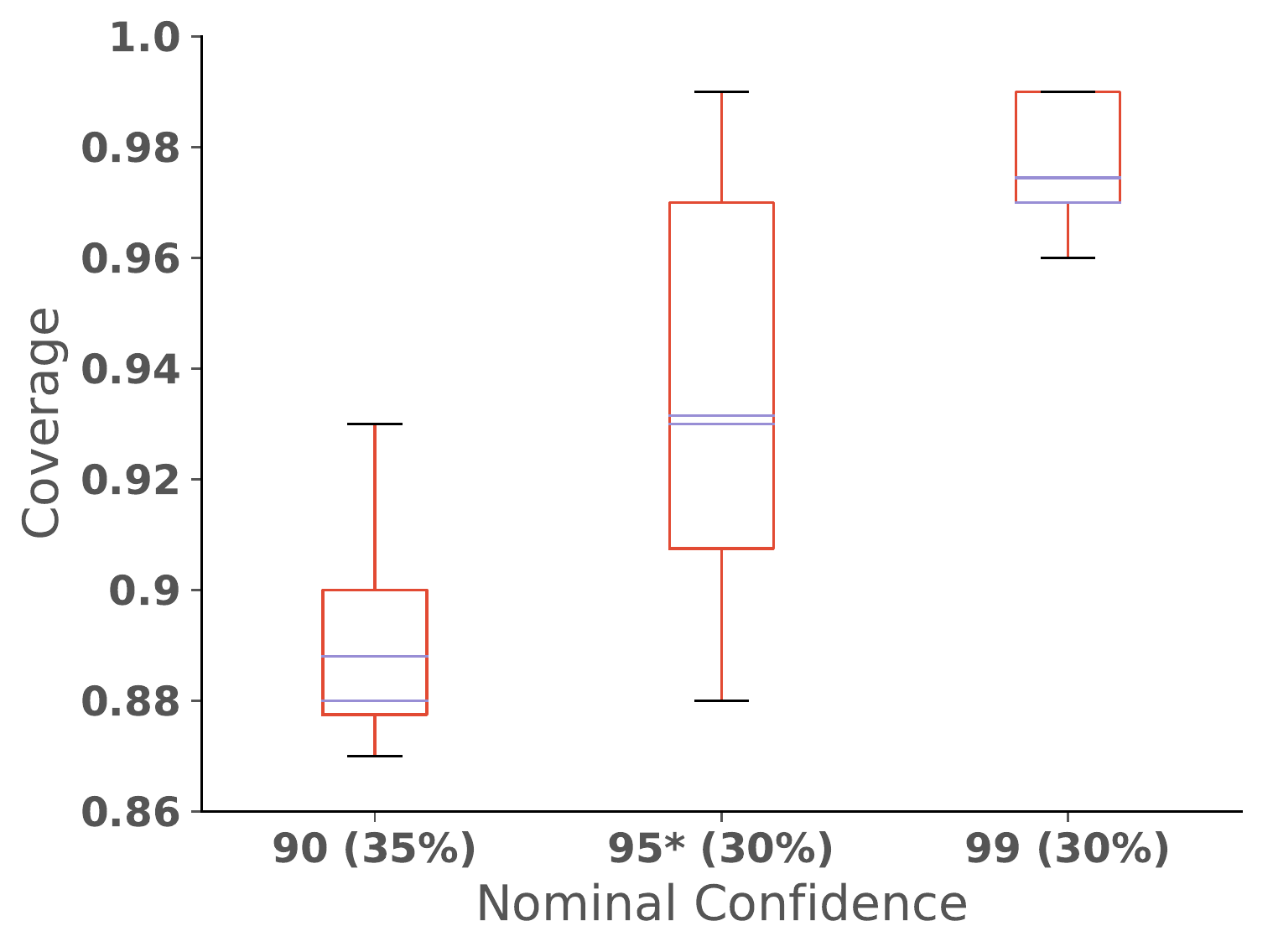} }}%
    \caption{Coverage distribution of the RF configurations in Keymind and Tukutuku.}%
    \label{fig:coverage}%
\end{figure}

In Figure~\ref{fig:coverage} (Keymind) we observe that 95\%, 100\%, and 0\% of configurations are reliable at NC90, NC95, and NC99, respectively. In Tukutuku, we observe that 35\%, 30\%, and 30\% of configurations are reliable at NC90, NC95, and NC99 respectively. However, the difference among configurations is statistically significant in only two out of six cases: Keymind NC90 and Tukutuku NC95. Therefore, in 33\% of the cases \textit{we can reject H\ref{nh:rq1_1_coverage}0: The coverage is equal among configurations}.

\begin{figure}[!t]%
    \centering
    \subfloat[Keymind]{{\includegraphics[width=.7\columnwidth]{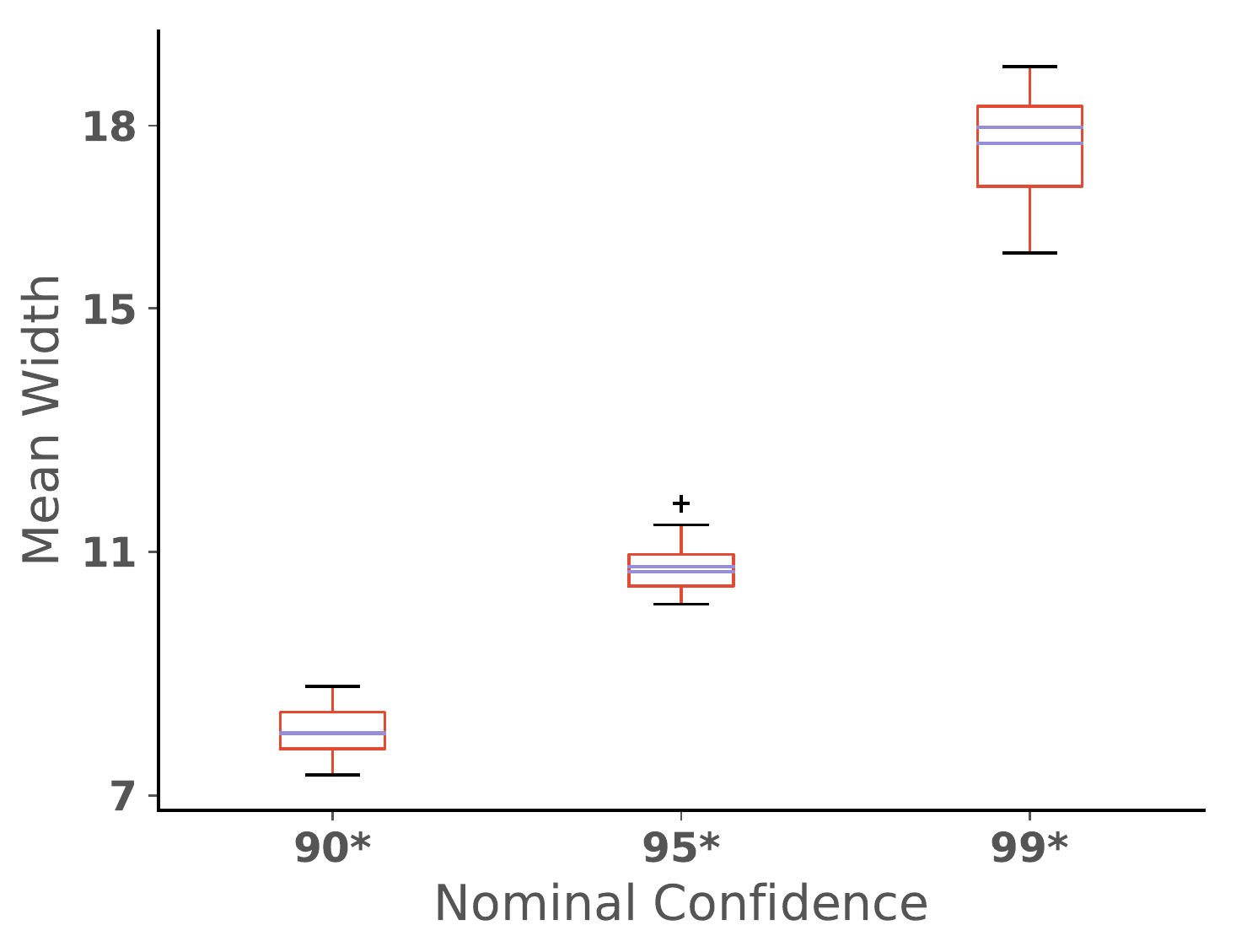} }}%
    \\
    \subfloat[Tukutuku]{{\includegraphics[width=.7\columnwidth]{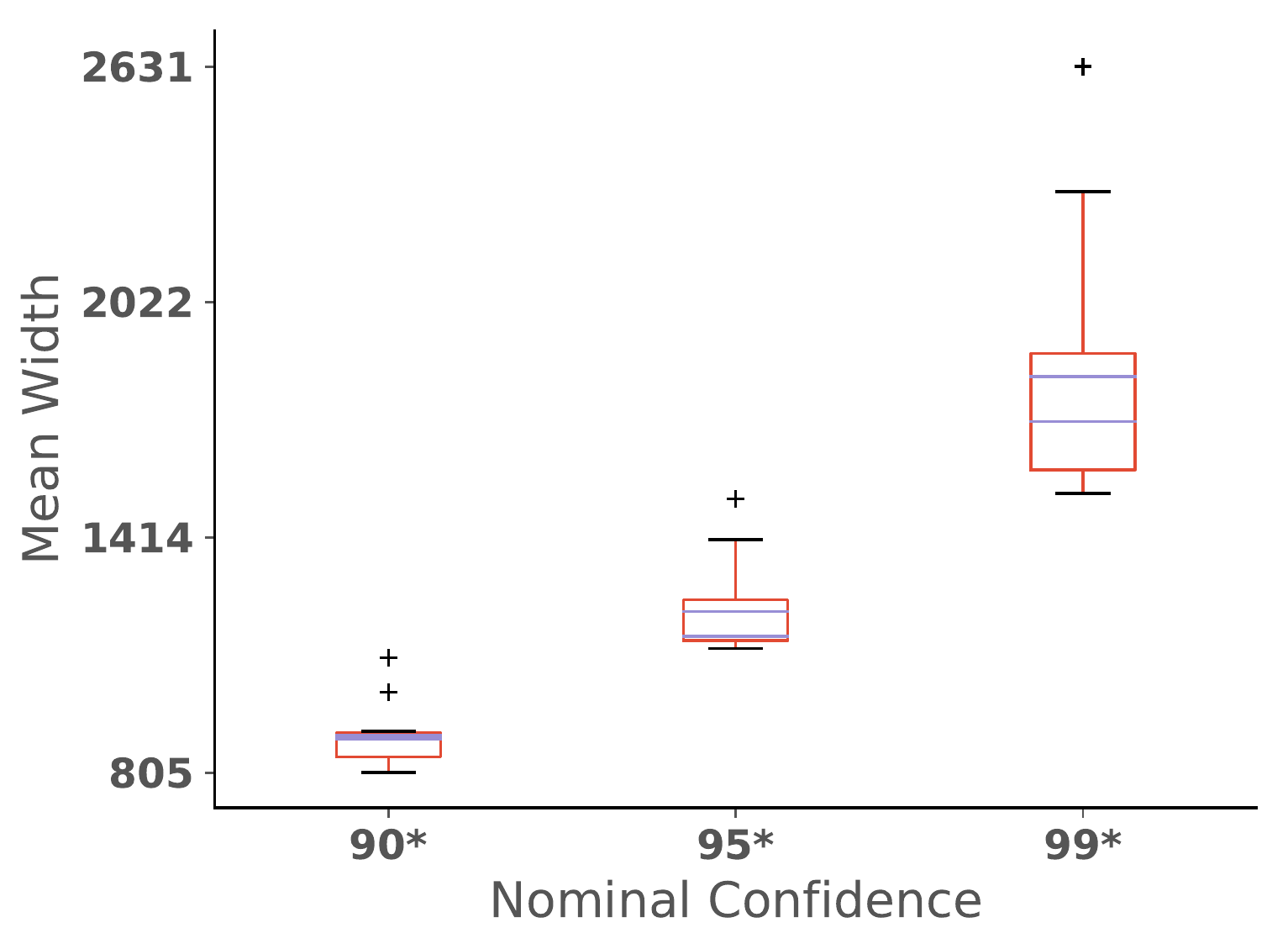} }}%
    \caption{Mean width distribution of the RF configurations in Keymind and Tukutuku.}%
    \label{fig:width}%
\end{figure}

Notably, in Figure~\ref{fig:coverage} (Tukutuku) we observe an overlap among coverages of different NCs; the configuration with the highest coverage at $NC_i$ has a higher coverage than the configuration with the lowest coverage at $NC_{i+1}$; \textit{choosing a particular configuration can increase coverage more than selecting a higher NC}. We note that it is not necessarily desirable to have a coverage much larger than the nominal confidence as this might imply that intervals are excessively wide. 

The Keymind dataset provides a useful context in which we can illustrate three situations.
\begin{enumerate}
    \item At NC90 there is one unreliable configuration. In this situation it is important to not choose this configuration.
    \item At NC95 we observe that all configurations are reliable. In this situation it is still important to choose among configurations in order to construct the narrowest PIs. 
    \item At NC99 we observe that all configurations are unreliable. In this situation, it is important to know that we cannot provide reliable prediction intervals.
\end{enumerate}
We reject \textit{H\ref{nh:rq1_1_width}0} in all six cases (see Figure~\ref{fig:width}). Thus, \textit{we are confident in rejecting H\ref{nh:rq1_1_width}0: The width is equal among configurations.}
By comparing Figure~\ref{fig:coverage} with Figure~\ref{fig:width} we observe that the distributions of coverage are larger than mean width. This suggests that configurations vary more with respect to coverage than width. However, statistical analysis supports the opposite conclusion. This could be due to the fact that in \textit{H\ref{nh:rq1_1_coverage}0} the variable is binary whereas in \textit{H\ref{nh:rq1_1_width}0} the variable is numeric. Hence, the two different tests can vary in statistical power, i.e., it could be that configurations vary more with respect to coverage than we are able to detect.

\begin{table}[!t]
\centering
\caption{Potential benefit of tuning in Keymind and Tukutuku.}
\label{tbl:tuning_potential_benefit}
\resizebox{.5\columnwidth}{!}{%
\begin{tabular}{|l|c|c|c|}
\Xhline{2\arrayrulewidth}
\textbf{Project} & \textbf{NC90} & \textbf{NC95} & \textbf{NC99} \\ \Xhline{2\arrayrulewidth}
Keymind & \cellcolor[HTML]{036400}{\color[HTML]{FFFFFF} \textbf{DU}} & \textbf{E} & \cellcolor[HTML]{00a700}{\color[HTML]{FFFFFF} \textbf{AU}} \\ \Xhline{2\arrayrulewidth}
Tukutuku & \cellcolor[HTML]{036400}{\color[HTML]{FFFFFF} \textbf{DU}} & \cellcolor[HTML]{036400}{\color[HTML]{FFFFFF} \textbf{DU}} & \cellcolor[HTML]{036400}{\color[HTML]{FFFFFF} \textbf{DU}} \\ \Xhline{2\arrayrulewidth}
\end{tabular}}
\end{table}

In Table~\ref{tbl:tuning_potential_benefit}, we observe that \textit{tuning is promising in five (83\%) cases}; i.e., there is only one case in which default is the best configuration (Keymind NC95). There are four cases (66\%) in which the default configuration is unreliable and there exists at least one reliable configuration. Lastly, the default configuration is unreliable in all cases in the Tukutuku dataset.

\begin{center}
\fbox{
\begin{minipage}[!t]{0.9\linewidth}
{\bf RQ1 Summary:}
In Tukutuku default is unreliable at all NCs. In Keymind default is best at only one NC. Thus, tuning is highly promising in both datasets. 
\end{minipage}
}
\end{center}


\subsection{RQ~\ref{rq:tune_benefit}: Are validation techniques beneficial?}\label{sec:rq1_2_results}
\subsubsection{Results}
Table~\ref{tbl:tuning_benefit}  presents the tags related to the benefit of applying a specific validation technique for each NC in both datasets. The table is formatted as follows: column 1 reports the project, column 2 reports the validation technique, and columns 3, 4, and 5 report the scenario tag at NC90, NC95, and NC99 respectively. To improve readability, we color coded the tags. Green indicates that a validation technique is beneficial; the darker the green, the higher its benefit. White indicates that a validation technique is neutral. Red indicates that a validation technique is counterproductive; the darker the red, the more it is counterproductive. For instance, looking at column 3 row 3 of Table~\ref{tbl:tuning_benefit} we can observe that at NC90 in Keymind the tag is DU, i.e., the default configuration is unreliable and the 25/75 validation technique is able to select a reliable configuration.

\begin{table}[!b]
\centering
\caption{Level of benefit provided in tuning via specific validation techniques in Keymind and Tukutuku}
\label{tbl:tuning_benefit}
\resizebox{.5\columnwidth}{!}{%
\begin{tabular}{|l|l|c|c|c|}
\Xhline{2\arrayrulewidth}
\textbf{Project} & \textbf{Technique} & \multicolumn{1}{l|}{\textbf{NC90}} & \multicolumn{1}{l|}{\textbf{NC95}} & \multicolumn{1}{l|}{\textbf{NC99}} \\ \Xhline{2\arrayrulewidth}
 & Bootstrap & \cellcolor[HTML]{00e400}{\color[HTML]{FFFFFF} \textbf{PU}} & \cellcolor[HTML]{680100}{\color[HTML]{FFFFFF} \textbf{TU}} & \cellcolor[HTML]{00a700}{\color[HTML]{FFFFFF} \textbf{APU}} \\ \cline{2-5} 
 & 10x10fold & \cellcolor[HTML]{00e400}{\color[HTML]{FFFFFF} \textbf{PU}} & \cellcolor[HTML]{CB0000}{\color[HTML]{FFFFFF} \textbf{SW}} & \cellcolor[HTML]{00a700}{\color[HTML]{FFFFFF} \textbf{APU}} \\ \cline{2-5} 
 & 25/75 & \cellcolor[HTML]{036400}{\color[HTML]{FFFFFF} \textbf{DU}} & \cellcolor[HTML]{CB0000}{\color[HTML]{FFFFFF} \textbf{SW}} & \cellcolor[HTML]{00a700}{\color[HTML]{FFFFFF} \textbf{APU}} \\ \cline{2-5} 
 & 50/50 & \cellcolor[HTML]{036400}{\color[HTML]{FFFFFF} \textbf{DU}} & \cellcolor[HTML]{680100}{\color[HTML]{FFFFFF} \textbf{TU}} & \cellcolor[HTML]{00a700}{\color[HTML]{FFFFFF} \textbf{APU}} \\ \cline{2-5} 
 & 75/25 & \cellcolor[HTML]{036400}{\color[HTML]{FFFFFF} \textbf{DU}} & {\color[HTML]{000000} \textbf{NSD}} & \cellcolor[HTML]{00a700}{\color[HTML]{FFFFFF} \textbf{APU}} \\ \cline{2-5} 
 & LOO & \cellcolor[HTML]{036400}{\color[HTML]{FFFFFF} \textbf{DU}} & \cellcolor[HTML]{CB0000}{\color[HTML]{FFFFFF} \textbf{SW}} & \cellcolor[HTML]{00a700}{\color[HTML]{FFFFFF} \textbf{APU}} \\ \cline{2-5} 
 & TSCV & \cellcolor[HTML]{F8A102}{\color[HTML]{FFFFFF} \textbf{NKU}} & \cellcolor[HTML]{CB0000}{\color[HTML]{FFFFFF} \textbf{SW}} & \cellcolor[HTML]{00a700}{\color[HTML]{FFFFFF} \textbf{APU}} \\ \cline{2-5} 
\multirow{-8}{*}{Keymind} & TSHVCV & \cellcolor[HTML]{036400}{\color[HTML]{FFFFFF} \textbf{DU}} & \cellcolor[HTML]{680100}{\color[HTML]{FFFFFF} \textbf{TU}} & \cellcolor[HTML]{00a700}{\color[HTML]{FFFFFF} \textbf{APU}} \\ \Xhline{2\arrayrulewidth}
 & Bootstrap & \cellcolor[HTML]{036400}{\color[HTML]{FFFFFF} \textbf{DU}} & \cellcolor[HTML]{00e400}{\color[HTML]{FFFFFF} \textbf{PU}} & \cellcolor[HTML]{00e400}{\color[HTML]{FFFFFF} \textbf{PU}} \\ \cline{2-5} 
 & 10x10fold & \cellcolor[HTML]{F8A102}{\color[HTML]{FFFFFF} \textbf{NKU}} & \cellcolor[HTML]{036400}{\color[HTML]{FFFFFF} \textbf{DU}} & \cellcolor[HTML]{00e400}{\color[HTML]{FFFFFF} \textbf{PU}} \\ \cline{2-5} 
 & 25/75 & \cellcolor[HTML]{00e400}{\color[HTML]{FFFFFF} \textbf{PU}} & \cellcolor[HTML]{00e400}{\color[HTML]{FFFFFF} \textbf{PU}} & \cellcolor[HTML]{00e400}{\color[HTML]{FFFFFF} \textbf{PU}} \\ \cline{2-5} 
 & 50/50 & \cellcolor[HTML]{036400}{\color[HTML]{FFFFFF} \textbf{DU}} & \cellcolor[HTML]{036400}{\color[HTML]{FFFFFF} \textbf{DU}} & \cellcolor[HTML]{00e400}{\color[HTML]{FFFFFF} \textbf{PU}} \\ \cline{2-5} 
 & 75/25 & \cellcolor[HTML]{F8A102}{\color[HTML]{FFFFFF} \textbf{NKU}} & \cellcolor[HTML]{00e400}{\color[HTML]{FFFFFF} \textbf{PU}} & \cellcolor[HTML]{00e400}{\color[HTML]{FFFFFF} \textbf{PU}} \\ \cline{2-5} 
 & LOO & \cellcolor[HTML]{F8A102}{\color[HTML]{FFFFFF} \textbf{NKU}} & \cellcolor[HTML]{F8A102}{\color[HTML]{FFFFFF} \textbf{NKU}} & \cellcolor[HTML]{00e400}{\color[HTML]{FFFFFF} \textbf{PU}} \\ \cline{2-5} 
 & TSCV & \cellcolor[HTML]{00e400}{\color[HTML]{FFFFFF} \textbf{PU}} & \cellcolor[HTML]{00e400}{\color[HTML]{FFFFFF} \textbf{PU}} & \cellcolor[HTML]{00e400}{\color[HTML]{FFFFFF} \textbf{PU}} \\ \cline{2-5} 
\multirow{-8}{*}{Tukutuku} & TSHVCV & \cellcolor[HTML]{036400}{\color[HTML]{FFFFFF} \textbf{DU}} & \cellcolor[HTML]{036400}{\color[HTML]{FFFFFF} \textbf{DU}} & \cellcolor[HTML]{00e400}{\color[HTML]{FFFFFF} \textbf{PU}} \\ \Xhline{2\arrayrulewidth}
\end{tabular}}
\end{table}

\subsubsection{Discussion}
The measurement design of this research questions consists of two datasets, three nominal confidences, and eight validation techniques, i.e., a total of 48 cases. We observe that tuning is \textit{beneficial} in 35 (73\%) cases, \textit{neutral} in one (2\%) case, and \textit{counterproductive} in 12 (25\%) cases. 

We note that no validation technique is beneficial at Keymind NC95. One possible explanation for this is that the default configuration is the best configuration (see Table~\ref{tbl:tuning_potential_benefit}), i.e., selecting anything other than default is counterproductive. 

All validation techniques are beneficial at NC99 in both Keymind and Tukutuku, specifically they deem all configurations as unreliable and default is in fact unreliable. However, 6 (30\%) configurations are reliable in Tukutuku (see Figure~\ref{fig:coverage} but no validation technique is able to select any of these. This result suggests \textit{there is room for improvement in the development of validation techniques.} 

Regarding the benefit in tuning with specific validation techniques, we note that holdout techniques (e.g. \textit{75/25}) are typically dismissed for being unreliable in performance estimation \cite{tanti2017empirical}. However, we find that \textit{75/25} selects a configuration with performance \textit{at least} as good as default, whereas \textit{bootstrapping}, \textit{10x10fold}, and \textit{TSCV} all select configurations with performance worse than default in one or more cases. Therefore, we observe that the \textit{75/25} holdout is, practically speaking, more beneficial within our datasets.  

Lastly, our results suggest that \textit{techniques which preserve order outperform techniques which do not preserve order}. Specifically, we observe that \textit{bootstrap, 10x10fold, and LOO are among the four least beneficial techniques}. This is an expected result with respect to software engineering because both datasets involve some notion of time; techniques that do not preserve order are likely inaccurate.

\begin{center}
\fbox{
\begin{minipage}[t]{0.9\linewidth}
{\bf RQ2 Summary:}
 No technique is beneficial in all NCs of both datasets. Moreover, most of the techniques are counterproductive in most of the cases. Thus, it is important to choose the right validation technique to avoid counterproductive effects of tuning. 
\end{minipage}
}
\end{center}

\subsection{RQ3: Are meta-validation technique beneficial?}\label{sec:rq2_2_results}
\subsubsection{Results}
Table~\ref{tbl:metatuning_benefit} presents the level of benefit of applying a specific meta-validation technique in a specific NC and  datasets. The table follows the same format as Table~\ref{tbl:tuning_benefit}.

\begin{table}[!b]
\centering
\caption{Meta-technique Benefit}
\label{tbl:metatuning_benefit}
\resizebox{.6\columnwidth}{!}{%
\begin{tabular}{|l|l|c|c|c|}
\Xhline{2\arrayrulewidth}
\textbf{Project} & \textbf{Technique} & \multicolumn{1}{l|}{\textbf{NC90}} & \multicolumn{1}{l|}{\textbf{NC95}} & \multicolumn{1}{l|}{\textbf{NC99}} \\ \Xhline{2\arrayrulewidth}
\multirow{3}{*}{Keymind} & Meta - 25/75 & \cellcolor[HTML]{036400}{\color[HTML]{FFFFFF} \textbf{DU}} & \cellcolor[HTML]{FE0000}{\color[HTML]{FFFFFF} \textbf{SW}} & \cellcolor[HTML]{00a700}{\color[HTML]{FFFFFF} \textbf{APU}} \\ \cline{2-4}
 & Meta - 50/50 & \cellcolor[HTML]{036400}{\color[HTML]{FFFFFF} \textbf{DU}} & \cellcolor[HTML]{680100}{\color[HTML]{FFFFFF} \textbf{TU}} & \cellcolor[HTML]{00a700}{\color[HTML]{FFFFFF} \textbf{APU}} \\ \cline{2-4}
 & Meta - 75/25 & \cellcolor[HTML]{036400}{\color[HTML]{FFFFFF} \textbf{DU}} & \textbf{NSD} & \cellcolor[HTML]{00a700}{\color[HTML]{FFFFFF} \textbf{APU}} \\ \Xhline{2\arrayrulewidth}
\multirow{3}{*}{Tukutuku} & Meta - 25/75 & \cellcolor[HTML]{036400}{\color[HTML]{FFFFFF} \textbf{DU}} & \cellcolor[HTML]{00e400}{\color[HTML]{FFFFFF} \textbf{PU}} & \cellcolor[HTML]{00e400}{\color[HTML]{FFFFFF} \textbf{PU}} \\ \cline{2-4}
 & Meta - 50/50 & \cellcolor[HTML]{F8A102}{\color[HTML]{FFFFFF} \textbf{NKU}} & \cellcolor[HTML]{F8A102}{\color[HTML]{FFFFFF} \textbf{NKU}} & \cellcolor[HTML]{00e400}{\color[HTML]{FFFFFF} \textbf{PU}} \\ \cline{2-4}
 & Meta - 75/25 & \cellcolor[HTML]{036400}{\color[HTML]{FFFFFF} \textbf{DU}} & \cellcolor[HTML]{00e400}{\color[HTML]{FFFFFF} \textbf{PU}} & \cellcolor[HTML]{00e400}{\color[HTML]{FFFFFF} \textbf{PU}} \\ \Xhline{2\arrayrulewidth}
\end{tabular}}
\end{table}
\subsubsection{Discussion}
The most important result of Table~\ref{tbl:metatuning_benefit} is that \textit{meta 75/25 holdout} is not counterproductive at any NC in either dataset; therefore, from a practical perspective, we observe that \textit{the use of meta-validation is more beneficial than using any validation technique across multiple contexts}.
\textit{Meta 75/25} is perfect at all NCs in Keymind, i.e., it chooses the validation technique providing the highest level of benefit at each of the three NCs. However, in Tukutuku, \textit{meta 75/25} is perfect at NC90 and NC99 but not at NC95. Specifically, \textit{meta 75/25} chooses the validation technique(s) providing the highest level of benefit at NC90 and NC99 in Tukutuku. However, at NC95 in Tukutuku, \textit{50/50} and \textit{TSHVCV} achieve a level of benefit of \textit{DU} (see Table~\ref{tbl:tuning_benefit}) whereas \textit{meta 75/25} achieves \textit{PU}. This means that \textit{meta 75/25} does not select \textit{50/50} or \textit{TSHVCV}. This suggests \textit{there is room for improvement in the development of meta-validation techniques}.

\begin{center}
\fbox{
\begin{minipage}[t]{0.9\linewidth}
{\bf RQ3 Summary:}
 Meta 75/25 is able to choose a validation technique that is beneficial in all NCs of both datasets. Thus, the use of a meta-technique avoids counterproductive effects of tuning. 
\end{minipage}
}
\end{center}

\section{Usability and Replicability}\label{sec:replicability}
We have the right to use but not to share the Keymind and the Tukutuku datasets. Thus, we cannot allow exact replication of our study. However, in order to enhance the study usability and replicability  we developed and shared \textit{Meta\_tune}\footnote{https://github.com/smbayley/meta_tune}, an open-source Python package for constructing and analyzing prediction intervals. In order to support the use of \textit{Meta\_tune}, we provide a README\textsuperscript{5} and a demo\footnote{https://youtu.be/jer9mpcZCuo}.

\textit{Meta\_tune} currently provides five analysis operations:
\begin{enumerate}
    \item Configuration coverage box-plots
    \item Configuration width box-plots
    \item Technique F1 and EMMRthe theE
    \item Tuning Benefit
    \item Meta-tuning Benefit
\end{enumerate}

To support replicability we tried \textit{Meta\_tune} on a very large open-source project, Apache Ant, as designed and collected by  \citet{Jureczko:2010} and available online\footnote{https://doi.org/10.5281/zenodo.268440}. Tables~\ref{tbl:ant} and \ref{tbl:meta_ant} report the results of \textit{Meta\_tune} on Apache Ant. The README and the demo provide step-by-step instructions for running \textit{Meta\_tune} on Apache Ant and get results in Tables~\ref{tbl:ant} and \ref{tbl:meta_ant}.

We note that results on Tables~\ref{tbl:ant} and \ref{tbl:meta_ant} are meant to be replicable but not generalize given that the data might not be designed and collected to support PIs.

\begin{table}[!h]
\centering
\caption{validation technique benefit in Ant}
\label{tbl:ant}
\resizebox{.6\columnwidth}{!}{%
\begin{tabular}{|l|c|c|c|}
\hline
\textbf{Technique} & \textbf{NC90} & \textbf{NC95} & \textbf{NC99} \\ \hline
100\_OOS\_Bootstrap & NSD & NSD & \cellcolor[HTML]{036400}{\color[HTML]{FFFFFF} \textbf{DU}} \\ \hline
10x10fold & NSD & NSD & \cellcolor[HTML]{F8A102}{\color[HTML]{FFFFFF} \textbf{NKU}} \\ \hline
25-75 & NSD & NSD & \cellcolor[HTML]{036400}{\color[HTML]{FFFFFF} \textbf{DU}} \\ \hline
50-50 & NSD & NSD & \cellcolor[HTML]{036400}{\color[HTML]{FFFFFF} \textbf{DU}} \\ \hline
75-25 & NSD & NSD & \cellcolor[HTML]{036400}{\color[HTML]{FFFFFF} \textbf{DU}} \\ \hline
LOO & NSD & NSD & \cellcolor[HTML]{F8A102}{\color[HTML]{FFFFFF} \textbf{NKU}} \\ \hline
TSCV & NSD & NSD & \cellcolor[HTML]{036400}{\color[HTML]{FFFFFF} \textbf{DU}} \\ \hline
TSHVCV & NSD & NSD & \cellcolor[HTML]{036400}{\color[HTML]{FFFFFF} \textbf{DU}} \\ \hline
\end{tabular}}
\end{table}

\begin{table}[!h]
\centering
\caption{Meta-75/25 benefit in Ant}
\label{tbl:meta_ant}
\resizebox{.6\columnwidth}{!}{%
\begin{tabular}{|l|c|c|c|}
\hline
\textbf{Technique} & \textbf{NC90} & \textbf{NC95} & \textbf{NC99} \\ \hline
Meta-25/75 & NSD & NSD & \cellcolor[HTML]{036400}{\color[HTML]{FFFFFF} \textbf{DU}} \\ \hline
Meta-50/50 & NSD & NSD & \cellcolor[HTML]{036400}{\color[HTML]{FFFFFF} \textbf{DU}} \\ \hline
Meta-75/25 & NSD & NSD & \cellcolor[HTML]{036400}{\color[HTML]{FFFFFF} \textbf{DU}} \\ \hline
\end{tabular}}
\end{table}

\section{Threats to Validity}\label{sec:threats}
In this section, we report the threats to validity related to our study. The description is organized by threat type, i.e., Conclusion, Internal, Construct, and External.

\subsection{Conclusion}
Conclusion validity regards issues that affect the ability to draw accurate conclusions about relations between the treatments and the outcome of an experiment \cite{Wohlin:2012}.

We tested all hypotheses with nonparametric tests (e.g., Friedman) which are prone to type-2 error, i.e,. not rejecting a false hypothesis. We have been able to reject the hypotheses in most of the cases; therefore, the likelihood of a type-2 error is low. Moreover, the alternative would have been using parametric tests (e.g., ANOVA) which are prone to type-1 error, i.e., rejecting a true hypothesis, which in our context is less desirable than type-2 error.

\subsection{Internal}\label{sec:threats_internal}
Internal validity regards the influences that can affect the independent variables with respect to causality \cite{Wohlin:2012}.

Results are related to the specific set of predictors used. We cannot check if these predictors are perfect. It could be that the set of predictors in use impacts the benefit of validation or meta-validation. We tried our best when developing the dataset (see Section~\ref{sec:design_keymind} and \citet{mendes2005investigating}) to include all predictors that are reasonably correlated with number of post-defects. Moreover, the use of RF minimizes this threat because it performs feature selection intrinsically.

\citet{7725218} report on the the time-space continuum threats to validity in validating recommendation systems based on past data. We believe this threat is low in the Keymind case since a prediction model was already in use when the releases happened.

\subsection{Construct}
Construct validity regards the ability to generalize the results of an experiment to the theory behind the experiment \cite{Wohlin:2012}.

The use of three specific coverage levels (i.e., NC90, NC95 and NC99) might influence our results. However we chose these coverage levels according to  \citet{thelin2002confidence} and the feedback of the analytic users, i.e., the Keymind developers and project managers. Specifically, it does not make sense to spend resources to develop a tool that is desired to be accurate less than 90\% of the time. Therefore, we did not consider nominal confidences less than 90\%. Lastly, a 100\% nominal confidence would have had increased the mean width by 55\% compared to a 99\% confidence (i.e., 17.7 vs 27.5), thus highly reducing the analytic actionability; therefore, we did not consider NC100. 

Configuration goodness is dependent on context-specific goal(s). Certain organizations might prioritize narrow intervals over reliable intervals.  In this paper, we had a specific goal in mind (i.e., to support the claim ``I am NC\% confident that the number of defects is between $a$ and $b$''). We recognize that our results might not generalize to organizations with different goals, even if we do not know of any such organizations. 

Moreover, the definition of the scenario tags as beneficial, neutral and counterproductive could be subjective. We defined and discussed the scenarios tags to pragmatically represent the desire of the analytic users, i.e., the Keymind developers and project managers.

Some might argue that a validation technique is beneficial if it selects the best configuration rather than a configuration that is better than the default. However, understanding if a validation technique selects the best configuration is interesting to know if the question to answer is if a validation technique is perfect rather than beneficial. Anyway, we show that no validation technique and meta-validation technique is perfect and this paves the way to future research effort. 

The default configuration might vary among implementations. The default MTRY for Sci-Kit Learn's RF regressor is 1.0. However, the default MTRY in R's\footnote{http://lojze.lugos.si/~darja/software/r/library/randomForest/html/randomForest.html} implementation is $\frac{n}{3}$, where $n$ is the number of predictors. The default MTRY in WEKA's implementation is $log_2(n) + 1$ where n is the number of predictors. It is possible that the benefit reported for tuning and meta-validation techniques varies among different default configurations. 

\citet{mayr2012prediction} suggest that the only way to demonstrate PI correctness is with an empirical evaluation based on simulated data (i.e., conditional coverage vs sample coverage). If true, this could limit the generalizability of our results. However, previous successful SE studies working with PIs have reported sample coverage \cite{jorgensen2004better, angelis2000simulation} rather than conditional coverage, and thus our results fit into existing SE research. 

\subsection{External}
External validity regards the extent to which the research elements (subjects, artifacts, etc.) are representative of actual elements \cite{Wohlin:2012}.

This study used only two datasets and hence could be deemed of low generalization compared to studies using tens or hundreds of datasets. However, as stated by \citet{nagappan2013diversity}, \textit{``more is not necessarily better.''} We preferred to test our hypotheses on datasets in which we were confident quality is high and that are close to industry. Moreover, in order to encourage replicability, we have provided \textit{meta-tune}, an open source python package for using and tuning PIs. 

Both Keymind and Tukutuku datasets are relatively large (363 and 195 industrial data points respectively). More common industry datasets could be smaller. Thus, we plan to investigate how the size of the dataset influences tuning and meta-tuning accuracy.

Neither Keymind nor Tukutuku are open-source projects. Thus, our results might not be generalizable to the context of open-source development. 

The benefit of tuning or meta-validation techniques depends on the goodness of the default configuration, i.e., if default is optimal there is no need to tune or meta-tune. Thus, the observed tuning and meta-tuning benefits might not generalize to datasets in which the default configuration is more frequently optimal. However, \citet{provost2000machine} and \citet{tosun2009reducing} suggest the default configuration is suboptimal in many cases. Further, even if default is the best configuration, this study is valuable as it demonstrates meta-tuning can be used to minimize the potential negative impact of tuning. 

\section{Conclusion}\label{sec:conclusion}
There are types of variables, such as those in defect and effort prediction, in which prediction intervals can provide more informative and actionable results than point-estimates. The aim of this paper was to investigate the use and optimization of prediction intervals by automatically configuring Random Forest.

We have been inspired by the recent positive results of tuning in software engineering and in Random Forest. As detailed in \citet{briand2017generalize}, software engineering research made in collaboration with industry, such as this work, ``does not attempt to frame a general problem and devise universal solutions, but rather makes clear working assumptions, given a precise context, and relies on trade-offs that make sense in such a context to achieve practicality and scalability.'' As is the case with any software engineering advance in industry, the presented research is impacted by human, domain, and organizational factors. To support the creation of a body of knowledge on tuning and meta-validation, we provided a Python package for tuning and meta-validation of prediction intervals.

We tune Random Forest by performing an exhaustive search search with a specific validation technique on a single Random Forest parameter since this is the only parameter that is expected to impact prediction intervals. This paper investigates which, out of eight validation techniques, are beneficial for tuning, i.e., which automatically choose a Random Forest configuration constructing prediction intervals that are reliable and with a smaller width than the default configuration. Additionally, we present and validate three meta-validation techniques to determine which are beneficial, i.e., those which automatically chose a beneficial validation technique.
Our validation uses data from our industrial partner (Keymind Inc.) and the Tukutuku Research Project, containing 363 and 195 industrial data points related to post-release defect prediction and Web application effort estimation respectively. We focus validation on three nominal confidence levels (i.e., 90\%, 95\% and 99\%), thus leading to six cases (three nominal confidences for each dataset). Results show that: 1) The default configuration is unreliable in five cases. 2) Previously successfully adopted validation techniques for tuning, such as 50/50 holdout and bootstrap, fail to be beneficial and are counterproductive in at least one case. 3) No single validation technique is always beneficial for tuning. 4) Most validation techniques are counterproductive at 95\% confidence level. 5) The \textit{meta} 75/25 holdout technique selects the validation technique(s) that are beneficial in all  cases. We note that these results are not meant to generalize to the entire machine learning domain. Rather, these results are specific to two industrial datasets related to defect prediction and effort estimation.

To our knowledge, this is the first study to construct prediction intervals using Random Forest on software engineering data. Further, this is the first study to investigate meta-validation in the software engineering domain. As such, there is significant room for future research. From a researcher's perspective, there is room for improvement in the following areas. 

\vspace{\baselineskip}\noindent 1) Random Forest prediction intervals, as there are cases where no configuration is reliable.
\begin{itemize}
    \item Distribution as inputs: we envision a model that accepts, as an input, a distribution of values, rather than a point-value \cite{sabetzadeh2013goal}.
    \item Adjusting intervals: we envision a mechanism, such as the one presented by \citet{kabaila2004adjustment}, measuring the error of prediction intervals during training and using this measurements to adjust the intervals produced during testing.
    \item Heterogeneous forests: we would like to investigate the accuracy of a Random Forest in which the underlying decision trees have different configurations.
    \item Combining human and model estimates: \citet{jorgensen2002combination} show that human-estimates can be used to construct informative PIs. We would like to incorporate human-based estimates into our construction of prediction intervals.
\end{itemize}

\noindent 2) Validation and meta-validation techniques, as there are configurations that are better than the selected ones. 
\begin{itemize}
    \item Better tuning and meta-validation techniques: this study evaluated eight validation techniques and three meta-validation techniques. These are large numbers considering no previous study (that we know of) applied more than one validation technique and no meta-validation techniques. However, there are techniques that we did not consider because they have not proven to be as successful, such as the non-repeated k-fold cross-validation.
    \item Meta-meta-validation: because meta-validation improved on tuning, it is interesting to investigate if meta-meta-validation improves on meta-validation, i.e., if we can automatically choose the meta-validation technique. However, the higher the level of meta evaluations, the smaller the training and testing dataset becomes, and hence the results will tend to become less generalizable. In other words, there is a point where ``more meta'' becomes counterproductive. We would like to investigate ``what is the appropriate level of meta?''
    \item Impact of size and noise: Several studies investigate how size and noise impact defect classification accuracy \cite{kim2011dealing, herzig2013s,rahman2013sample}. On the same vein, we plan to investigate how size and noise impact the construction of narrow and reliable prediction intervals.
    \item Semi-automated decisions: on the one hand, we like to make decisions based on historical data rather than subjective opinions. In this study, we used tuning for making decision about the configuration to use, and used meta-validation for making decisions about the validation technique to use. On the other hand, data is rarely perfectly accurate or complete. Therefore, we need to develop a body of knowledge to understand when a parameter or a technique is likely to be better than another. 
\end{itemize}

From a practitioner's perspective, we found the Random Forest to be a reliable approach to construct prediction intervals. However, because the default configuration is frequently unreliable, the use of Random Forests must be accompanied with an accurate tuning technique. Widely adopted validation techniques in tuning such as the 50/50 holdout and bootstrap were frequently counterproductive. Thus, we recommend practitioners to perform meta-validation to minimize the possible counterproductive effects of tuning.

\bibliographystyle{plainnat}
\bibliography{bibliography} 

\section*{Appendix}\label{sec:appendix}

We divide the Keymind predictors in two groups: predictors having prior studies supporting their impact on defectiveness and predictors that are intuitively correlated with defectiveness. Tables~\ref{tbl:keymind_predictors1} and \ref{tbl:keymind_predictors2} present the names and information about each of the selected predictors.

\begin{table*}[t]
\centering
\caption{Keymind predictors with evidence of impacting defectiveness}
\label{tbl:keymind_predictors1}
\begin{tabular}{|p{.2\linewidth}|p{.3\linewidth}|p{.3\linewidth}|}
\hline
Predcitor & Definition & Rationale \\ \hline
Project Age & The duration, in days, from the start of the first release to the start of the current release. & More mature projects are less likely to be defect prone than new projects \cite{d2012evaluating}. \\ \hline
Number of Developers & The number of unique developers that made a SVN commit during the release. & Releases with more developers are more likely to be defect prone \cite{d2012evaluating}. \\ \hline
Number of New Developers & The number of unique developers that made a SVN commit during the current release and had not made a commit during any previous release. & Releases with more new developers are more likely to be defect prone \cite{d2012evaluating}. \\ \hline
Prior Number of God Classes & The number of god classes that were present in the project at the end of the previous release. & Projects with more god classes are more likely to be defect prone \cite{d2012evaluating}. \\ \hline
Project Architecture & A categorical feature that specifies the type of project (e.g., .NET standard vs MVC standard) & Some project architectures require more developers and/or code to complete a requirement. \\ \hline
\end{tabular}
\end{table*}

\begin{table*}[t]
\centering
\caption{Keymind predictors that are intuitively correlated with defectiveness}
\label{tbl:keymind_predictors2}
\begin{tabular}{|p{.2\linewidth}|p{.3\linewidth}|p{.3\linewidth}|}
\hline
Predictor & Definition & Intuition \\ \hline
Development Duration & The duration, in days, from the start of the releases to the end of the releases & Longer durations are indicative of releases with more requirements, and hence releases that are more defect prone. Moreover, shorter durations with a high number of features are more defect prone. \\ \hline
Estimated Effort & The total estimated effort, in hours, required for development. & More estimated effort is indicative of releases with more complex features, and hence releases that are more defect prone. \\ \hline
Testing Time & The total amount of time, in hours, spent on internal testing. & More time spent testing reveals defects prior to release, and hence makes releases less defect prone. \\ \hline
Number of Change Requests & The number of JIRA tasks that specify a change to the implementation of a previously specified requirement. & More change requests are indicative of releases with more development, and hence releases that are more defect prone. \\ \hline
Number of Implementation Tasks & The number of JIRA tasks that specify implementation of a new requirement. & More implementation tasks are indicative of releases with more development, and hence releases that are more defect prone. \\ \hline
Number of Release Candidates & The number of internal releases before the external release. & More release candidates is indicative of releases that have been exposed to more internal use, and hence releases that are less defect prone. \\ \hline
Number of Users & The number of end-users. & More end-users is indicative of releases that will be exposed to more external use, and hence releases that are more defect prone. \\ \hline
\end{tabular}
\end{table*}


\end{document}